\documentclass[letterpaper]{article} % DO NOT CHANGE THIS
\usepackage{aaai2026}  % DO NOT CHANGE THIS
\usepackage{times}  % DO NOT CHANGE THIS
\usepackage{helvet}  % DO NOT CHANGE THIS
\usepackage{courier}  % DO NOT CHANGE THIS
\usepackage[hyphens]{url}  % DO NOT CHANGE THIS
\usepackage{graphicx} % DO NOT CHANGE THIS
\usepackage{natbib}  % DO NOT CHANGE THIS AND DO NOT ADD ANY OPTIONS TO IT
\usepackage{caption} % DO NOT CHANGE THIS AND DO NOT ADD ANY OPTIONS TO IT
\usepackage{times}
\usepackage{soul}
\usepackage{url}
\usepackage[utf8]{inputenc}
\usepackage{amsmath}
\usepackage{amsthm}
\usepackage{booktabs}
\usepackage{algorithm}
\usepackage{algorithmic}
\usepackage{amssymb}

\usepackage{tabularx}
\usepackage{algorithm}
\usepackage{algorithmic}

\usepackage{newfloat}
\usepackage{listings}
\DeclareCaptionStyle{ruled}{labelfont=normalfont,labelsep=colon,strut=off} % DO NOT CHANGE THIS
\floatstyle{ruled}
\newfloat{listing}{tb}{lst}{}
\floatname{listing}{Listing}
\title{Do Judges Behave Like Algorithms?}
\author{
    Riya Manchanda\equalcontrib\textsuperscript{\rm 1},
    Eric Chen\equalcontrib\textsuperscript{\rm 1},
    Chloe Zhu\textsuperscript{\rm 1},
    Cynthia Rudin\textsuperscript{\rm 1},
    Brandon Garrett\textsuperscript{\rm 1},
    Songman Kang\textsuperscript{\rm 2}
}
\affiliations{
    \textsuperscript{\rm 1}Duke University\\
    \textsuperscript{\rm 2}Sungkyunkwan University\\
   riya.manchanda@duke.edu, eric.y.chen@duke.edu, qinyu.zhu@duke.edu, cynthia@cs.duke.edu, bgarrett@law.duke.edu, songmankang@skku.edu

}

\begin{document}

\maketitle

\begin{abstract}
What if judges already behave like algorithms? As artificial intelligence and algorithms are deployed in many settings, including the judicial system, many have debated whether judges should be allowed to rely on them. We suggest this debate about the role of algorithms in the justice system tracks a longstanding debate regarding the comparative benefits of judges relying on rules versus standards. Rules are clear, consistent, and predictable, but may not capture individual and equitable circumstances that more flexible standards rely on. We do not claim when or whether rules or standards are preferable. Instead, we ask whether judges follow predictable, algorithmic-like rules already. If judges already follow consistent, formula-like rules based on discrete and static factors such as criminal history, age, and charge type, then judicial behavior may be improved. However, if judges rely on individualized information that cannot be identified through court data, then standards-based decision-making may be more challenging to understand or improve. This work explores these questions by studying judicial decision-making in misdemeanor bail hearings in Harris County, Texas. Using available court data, we investigate whether magistrate judges follow what resembles an algorithm; whether they consider the same variables in their decision-making; and whether they are consistent with themselves and with each other. To do this, we train machine learning models for each judge, measure variable importance metrics to determine important variables for each judge's decision-making, and analyze outcomes of similar cases for judges. Our results reveal that these judges generally behave algorithmically: their decisions can be captured by small, interpretable formulas. However, in some cases, judges differ substantially, leading to surprising inconsistency and unequal treatment across similar defendants. Identifying cases where algorithms do not explain judicial decision-making can improve the justice system by focusing attention on decisions where individualized standards, rather than rules, better explains outcomes.
\end{abstract}
\begin{links}
    \link{Full Appendix and Public Code}{https://github.com/eychen2/AreJudgesAlgorithms}
\end{links}

\section{Introduction}

Having an impartial judge decide one’s dispute is an essential element of due process, or fair trial rights, around the world \citep{garrett2025,deeks2019,crootof2022}. An important dimension of due process is having a judicial officer who makes these decisions in a reasoned and consistent manner. Yet, there has been a longstanding tension between making decisions based on \textit{rules}, which can lead to consistent and predictable outcomes, and the use of more discretionary \textit{standards}, which can provide more equitable outcomes based on individual, case-specific considerations \citep{schlag1985}. There are important questions concerning when and whether rules versus standards are desirable, and there are tradeoffs concerning the choice of which to employ in legal decision-making \citep{kaplow1992}. For more information about rules and standards, refer to Appendix \ref{app:RulesStandards}.
In recent years, the debate about the use of algorithms and AI in the justice system has tracked many of the same concerns expressed in the longstanding debate between the comparative desirability of rules versus standards. One side argues that many algorithms are reductive and fail to account for the case-specific factors that a human judge must consider \citep{harcourt2015}. This concern is particularly pressing in areas where equity and a judge’s discretionary moral judgment are involved \citep{re2019}. Others note that human judges are also prone to certain types of inconsistencies and biases in their decision-making \citep{Rachlinkski2017,eaglin2023}. The type of data and the type of judicial task may also affect the choice of whether algorithms or discretionary judging are more desirable. Using AI or algorithms is more advantageous if prediction is straightforward and rule-based, while an element of human judgment will be more advantageous if omitted variables make accurate prediction more difficult \citep{fagan2019}. 
Highly relevant to this debate is whether judges \textit{already} behave like algorithms and follow rules, or rather, whether they tend to decide cases based on more case-specific standards. Let us define this question more carefully. By an ``algorithm,'' we refer to a simple formula, perhaps consisting of a few understandable rules. This simple formula involves only a few additions, subtractions, and multiplications. We call such formulas ``interpretable'' or ``understandable,'' because the actual factors and weights they rely upon are provided \citep{garrett2023}. We are not referring to ``black boxes,'' i.e., complicated or proprietary mathematical formulas, since we believe those should be avoided in criminal justice decision-making, as they raise serious constitutional concerns and have not been shown to be better than interpretable models \citep{garrett2023,garrettrudin2024}. The question we ask here is when and whether judges act like simple formulas. Are judges consistent in their judgments the same way a simple formula (or rule) would be? If so, can we identify these formulas or rules? If instead they employ standards, can we identify them in order to shed light on how they make their decisions? 
These questions are central to debates on whether algorithms should be used in the justice system. If judges already follow consistent, formula-like rules based on quantifiable cues such as prior offenses, age, and charge type, then judicial behavior could be understood and even improved through interpretable models rather than being replaced by them. Conversely, if their decisions deviate widely from any learnable rule, then the challenge is to identify these more flexible standards, assess the consistency of decision-making, and, in turn, ensure that the use of those standards can be understood and even improved. 

We examine this question by focusing on data from the third-largest urban county in the United States, Harris County, Texas, which includes the city of Houston. Our data is from the period after the 2019 entry of the O'Donnell Consent Decree, a landmark reform in Harris County designed to eliminate wealth-based disparities in bail decisions, and gave magistrates greater discretion in their cases, particularly regarding a smaller number of more serious misdemeanor cases.

In this work, we examine the decision-making by magistrates who are hired by the County to report to the elected judges and to make initial decisions regarding the comparatively more serious misdemeanor cases.  Specifically, we focused on the cases in which the magistrates decide, after a bail hearing, during which they hear from the district attorney and public defender, whether to recommend that a person be given a personal bond and released, perhaps with supervision conditions, or whether a person should be given a secured bond with a cash bail amount that they would have to pay to be released, possibly also with supervision conditions if they are able to pay or obtain a bond for that amount. Given the size of the jurisdiction, there are tens of thousands of these bail hearings and magistrate decisions each year. Our unique dataset contains over 22,000 unique cases decided by 21 magistrates.

To clarify, this work does not advocate replacing judges with algorithms. An essential concern is \textit{recourse:} humans are always needed, including when an algorithm is wrong, when cases raise individual or extenuating circumstances, or in cases where a rule-based decision would be appropriate, but the data about the individual is incorrect and the algorithm’s input needs to be adjusted \citep{garrett2023}. Regarding the latter situation, while cases with data errors should be rare, they are not in practice, partly due to pervasive errors in administrative criminal justice data \citep{pepper2009}. The first type of case -- where equitable discretion or standards should be used -- raises a normative question about the desirability of using equitable, standards-based approaches. Without addressing that normative question, we instead ask whether judges act like algorithms generally, including whether they are consistent with each other; whether they are needed mainly to intercept unusual cases; and whether there is some category of individualized decision-making that they already engage in. Examining the potentially algorithmic nature of judicial reasoning -- and the types of cases that judges do not view as suitable for rule-based approaches -- opens a path toward improved decision making and greater accountability.  \\
\noindent\textbf{Summary of experiments and results:}
To investigate our first question --  ``\textbf{When do judges not follow any reasonable algorithm, and are there circumstances leading to these inconsistencies?}'' -- we generated the ``Rashomon set'' of every possible good model of a flexible, nonparametric function class, namely, sparse decision trees. Each of these models represents a possible simple decision function that matches the judge’s decisions as accurately as possible. For each magistrate, we identified cases that were misclassified for every function in the Rashomon set; these ``unexplainable cases'' are judicial decisions where no good model can predict the judge’s ruling correctly. 

Unexplainable cases amount to 2\%-33\% of each judge’s cases. Manual analysis of a sample of hearing records and probable cause forms revealed that 100\% of the sample of these unexplainable cases arose from either: (1) data-entry or form errors, (2) information about charges outside Harris County that were not represented in the dataset, or (3) case-specific qualitative details that judges referenced but were absent from structured data. These findings suggest that even if much of judicial reasoning were algorithmic, real-world inconsistencies often stem from gaps in information and record-keeping rather than deliberate arbitrariness. These results suggest that there are systematic gaps in record-keeping that could be mitigated.

Our second question is \textbf{``Can each judge be described by a simple algorithm?''} We attempted to model each judge with a simple decision tree and reported the degree to which each magistrate's bail decisions can be represented by such rules (disregarding the unexplainable cases, which arise from a noisy generative process rather than a deliberate one). We found that across thousands of cases, most judges could be approximated by small decision trees (depth $\leq$5) using features such as prior revocations, pending cases, and defendant age, achieving accuracies between 75\% and 100\%, where at least 85\% of the decisions for each judge can be algorithmically determined for 17 of the 21 judges. This indicates that a substantial portion of judicial behavior reasonably follows systematic patterns that can be summarized in a few simple conditions, although some judges behaved more algorithmically than others. 

Digging further, we performed a variable importance analysis to answer: \textbf{``Do judges consider the same variables in decision making?''} Most variable importance analyses are flawed since they only consider variable importance for one model, despite the existence of the Rashomon set -- the set of good models for the data -- which often contains many equally accurate models that use completely different variables. Our analysis, instead, averages variable importance over good models to identify variables that are generally important. We find that there are three generally important variables across judges -- age, warrantct (number of previous warrants), and pre-defined flags of specific types of crime -- but the degree to which variables are important and the choice of additional important variables differ substantially between judges. Thus, even if each judge behaves algorithmically, they act like totally different algorithms. 

Finally, we delve deeper into the differences between judges, asking: \textbf{``Are judges consistent with each other? When do they disagree?''} To do so, we test whether a model trained to predict one judge’s decisions can accurately predict another judge’s decisions. We find that it cannot: judges exhibit meaningfully distinct decision-making processes. Specifically, when we evaluated each judge-specific model on cases decided by other judges, performance deteriorated substantially, revealing pronounced heterogeneity across judge pairs. We then identified structured ``disagreement rules'' that characterize when particular judges diverge. For instance, Judge 9 and Judge 4 disagree in 94\% of cases involving individuals aged 24 or younger, suggesting systematically different treatment of younger individuals. This indicates that the justice system’s outcomes depend heavily on who the judge is. A triplet consistency experiment, where we matched similar cases for different judges, confirmed these findings: judges were highly self-consistent, getting similar outcomes for similar cases ~76\%  of the time, but only modestly consistent with one another, coming to the same rulings on similar cases only $\sim$55\% of the time (close to random guessing), indicating that the justice system’s outcomes depend heavily on who the judge is.

Our results thus reveal a paradox. On one hand, judges do behave algorithmically a lot of the time; their decisions can be captured by small, interpretable formulas that rely on measurable variables. On the other hand, the algorithmic behavior differs from judge to judge, leading to inconsistency and unequal treatment across defendants. Identifying and comparing these implicit rule sets could enable data-driven judicial training, standardized guidelines, and mechanisms to promote interpretability and system-wide fairness.

By understanding when and why judges deviate from interpretable rules, we can better identify errors in data, biases in reasoning, and opportunities for reform. Ideally, with additional data on outcomes, we can focus judges on rules that produce desirable outcomes  \citep{loeffler2015}, while being a comparatively inexpensive type of intervention \citep{ludwig2024}.
Ultimately, we claim the central question is not whether algorithms belong in the justice system, but whether humans within it already act like them and whether we can make both better.

\section{Related Works}

A growing empirical literature studies judicial decision-making as a form of human judgment under uncertainty, particularly in high-volume and time-constrained settings such as pretrial bail hearings. This work consistently documents substantial variation across judges, even when they are presented with similar cases, raising concerns about consistency, equity, and arbitrariness in legal outcomes.

In the context of bail, several studies compare judicial decisions with algorithmic predictions. \citet{kleinberg2018human} analyze pretrial release decisions in New York City, showing that simple machine learning models can outperform judges in predicting failure to appear while simultaneously reducing incarceration and racial disparities. Other work \citep{arnold2022} demonstrates that judge-specific preferences contribute meaningfully to variation in bail outcomes, with some judges systematically imposing more restrictive conditions than others, even after controlling for observable case characteristics. These findings highlight the limits of unaided human judgment in repetitive, prediction-driven legal tasks.

These concerns are especially salient in Harris County, Texas, following the \textit{O'Donnell} Consent Decree. Before the decree, misdemeanor bail decisions were governed by a rigid bail schedule that tied release to the ability to pay, resulting in widespread detention of low-income defendants. Post-decree evaluations \citep{heaton2017} find that the reforms substantially reduced wealth-based detention, but did not eliminate heterogeneity in magistrate decision-making. In carve-out cases that proceed to individualized bail hearings, outcomes remain sensitive to which magistrate presides, suggesting that discretion continues to play a central role.

Beyond the legal literature, research in cognitive psychology provides insight into why judicial decisions may be well approximated by simple, interpretable rules. A large body of work shows that human experts frequently rely on heuristics—fast and frugal decision rules that trade optimality for cognitive efficiency. Foundational work on availability and representativeness heuristics \citep{Tversky1974} demonstrates how humans simplify probabilistic reasoning, while other works \citep{Marewski31032012}  show that in domains such as medical decision-making, heuristic-based models can rival more complex statistical approaches. These findings suggest that judicial reasoning, particularly under time pressure and informational constraints, may naturally take a rule-like or algorithmic form.

Consistent with this perspective, prior studies have found that expert decisions in high-stakes domains can often be captured by compact decision rules. Models of bail decisions in the United Kingdom using fast-and-frugal trees found that judges rely on a small number of cues applied in a consistent order \citep{dhami2001}. Similar results have been observed in other expert domains, including military and medical decision-making, where simple decision trees achieve accuracy comparable to more complex models while remaining interpretable.

Again, we do not ask whether algorithms should replace judges or whether algorithmic predictions outperform those of human decision-makers. Instead, we ask whether judges themselves already behave like algorithms: whether simple, interpretable rules can capture their decisions; whether those rules are consistent over time; and when no reasonable algorithm explains a judge’s behavior. By shifting the focus from human-versus-algorithm comparisons to the internal structure of judicial decision-making, our work connects normative debates about rules and standards with empirical evidence on human judgment, offering a new lens on accountability and consistency in the justice system.

\section{Data and Processing}
In Harris County, the Consent Decree gives the public access to certain data concerning misdemeanor pretrial release and detention decisions, including defendant characteristics (such as race, sex, and age), case characteristics (such as charges filed, booking and release dates), which magistrates and judges are assigned to cases, decisions made at bail hearings, and bail outcomes in cases from 2009 to the present. We focused on misdemeanor and felony cases from 2020 to 2025, a time period during which the Consent Decree rules had been adopted and were being used consistently. The dataset contained 531,921 unique cases, out of which we filtered out carve-out cases (cases that were not given a General Order Bond), duplicate cases, and off-docket cases (pre-screened cases where pretrial services had determined the defendant was eligible for release on a personal bond and where the magistrate agrees, granting the personal bond without a bail hearing). Using this data, we combined the crime case details and bond analytics information from the felony dataset and misdemeanor datasets for individual defendants to compute the features in Table \ref{tab:features} in Appendix \ref{app:datadesc}. These features mostly pertain to that defendant’s criminal history: one for past misdemeanors and one for past felonies (with the `\_$f$` suffix). Felony crime information is included because we discovered that magistrates sometimes grant personal bonds despite pending felony charges, reasoning that the ongoing case would ensure punishment if a reoffense occurred, making it important for modeling their decisions. A full description of the dataset and processing is in Appendix \ref{app:data}.

After feature engineering, we  excluded cases with any existing holds and deferred adjudications, since magistrates appeared to view these existing constraints as sufficient safeguards when granting personal bonds, as well as cases with null values for any of the features, leading to a total of 22,361 usable cases in our dataset. Out of these, 18,456 cases (82.5\%) had personal bond not granted (class 0) and 3,905 cases (17.5\%) had personal bond granted (class 1). 
\begin{figure}[t]
    \centering
    \includegraphics[width=.9\linewidth]{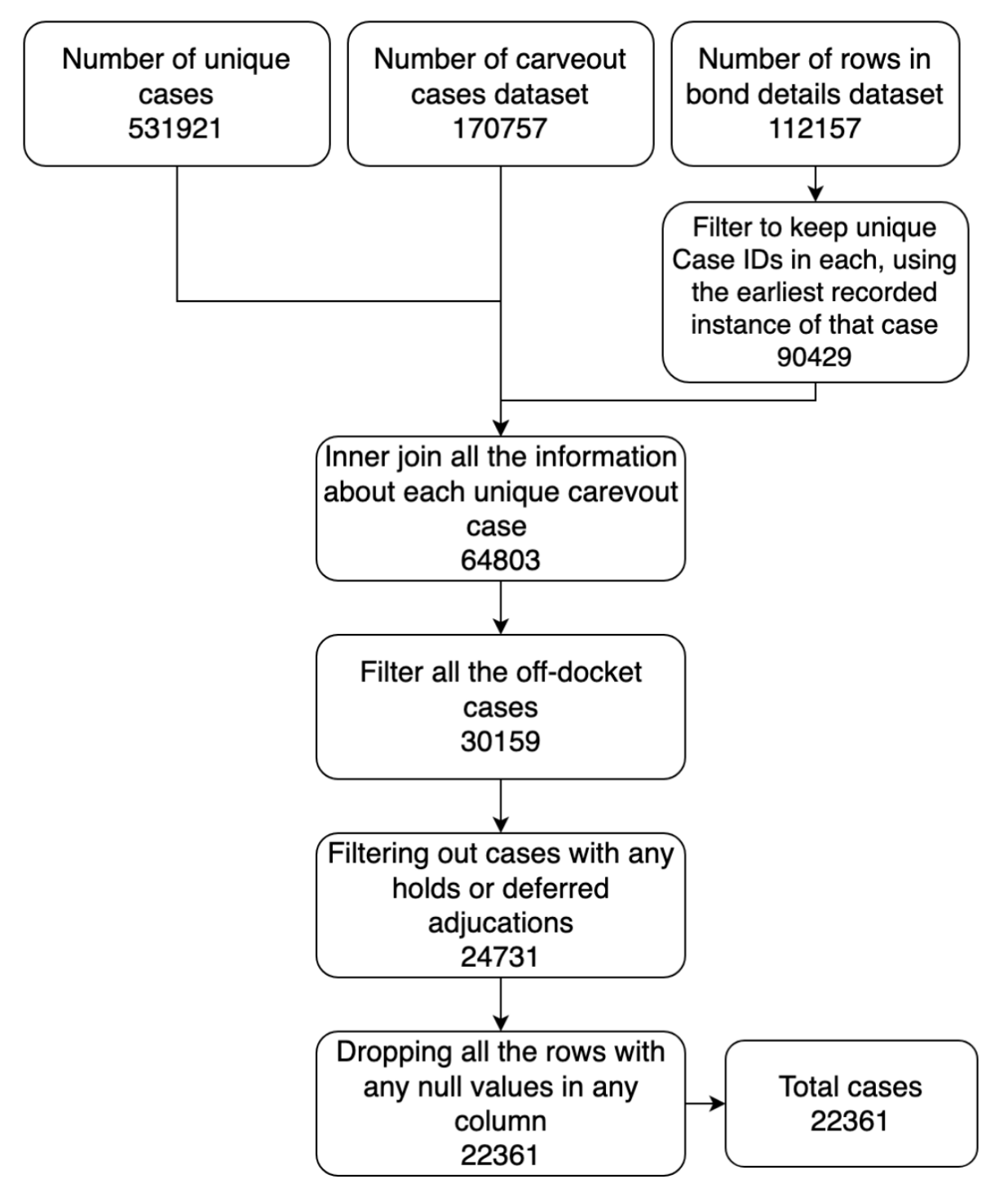}
    \caption{Data Processing  Workflow. The flowchart outlines steps for merging and cleaning judicial case records.}
    \label{fig:DataProcessing}
\end{figure}
Figure \ref{fig:DataProcessing} shows our data processing workflow.
After processing, we decided to use 21 of the 24 judges, excluding those who adjudicated too few cases to support reliable analysis. We assigned each included judge an anonymous identifier.

\section{Experiments and Results}
\begin{table*}[t]
\centering
\scriptsize
\setlength{\tabcolsep}{3pt}
\begin{tabular}{lccccccccccccccccccccc}
\toprule
\textbf{Judge} 
& 1 & 2 & 3 & 4 & 5 & 6 & 7 & 8 & 9 & 10 
& 11 & 12 & 13 & 14 & 15 & 16 & 17 & 18 & 19 & 20 & 21 \\
\midrule
\textbf{Rashomon Set Size} 
& 1188 & 612 & 682 & 40 & 272 & 354 & 76 & 24 & 7 & 16
& 852 & 210 & 1164 & 976 & 62 & 338 & 117 & 513 & 388 & 28 & 1088 \\
\textbf{Best Model Accuracy} 
& 78\% & 70\% & 73\% & 66\% & 73\% & 79\% & 66\% & 84\% & 85\% & 99\%
& 61\% & 83\% & 78\% & 69\% & 98\% & 83\% & 65\% & 78\% & 83\% & 85\% & 80\% \\
\bottomrule
\end{tabular}
\caption{Summary of Rashomon set size and representative model accuracy across judges.}
\label{tab:rashomon-judge}
\end{table*}
We structure our experiments around our four guiding research questions.
\subsection{When do judges not follow any reasonable algorithm?}
%Are there circumstances leading to these inconsistencies?}

Here, we find decisions that are not consistent with any reasonable algorithm characterizing the decision-making process of the judge, based on the available court data. We call these ``unexplainable cases'' since these decisions are seemingly incompatible with the judge’s usual way of reasoning. Such cases could arise from a variety of situations: evidence not in the database (missing data), incorrect data, personal bias of the judge, corruption, or randomness in decisions.

To find such cases, we used the TreeFARMS algorithm \citep{xin2022} to obtain a \textit{Rashomon set} of interpretable trees for each judge. A Rashomon set is the collection of models from a given function class (here, simple decision trees) that are the most performant on the dataset. Whereas standard machine learning methods provide one ``best'' model, we do not, because there can be many models that perform about equally well, and those models differ from each other \citep{SemenovaRuPa2022,SemenovaEtAl2023, FisherRuDo19, RudinEtAlAmazing2024}. 
TreeFARMS  is designed to efficiently enumerate and represent the Rashomon set of sparse decision trees \citep{xin2022} for a given dataset. Using TreeFARMS allows us to study variation across equally good models, examine which features consistently appear as important, and highlight areas of uncertainty or flexibility in decision-making.
Details about generating the Rashomon sets are in Appendix \ref{app:Rashomon}.
Table \ref{tab:rashomon-judge} shows the results of the Rashomon set generation, including the number of trees for each judge and best model accuracies, ranging from 60\% to 99\%. 

\begin{figure}[htbp]
    \centering
    \includegraphics[width=\linewidth]{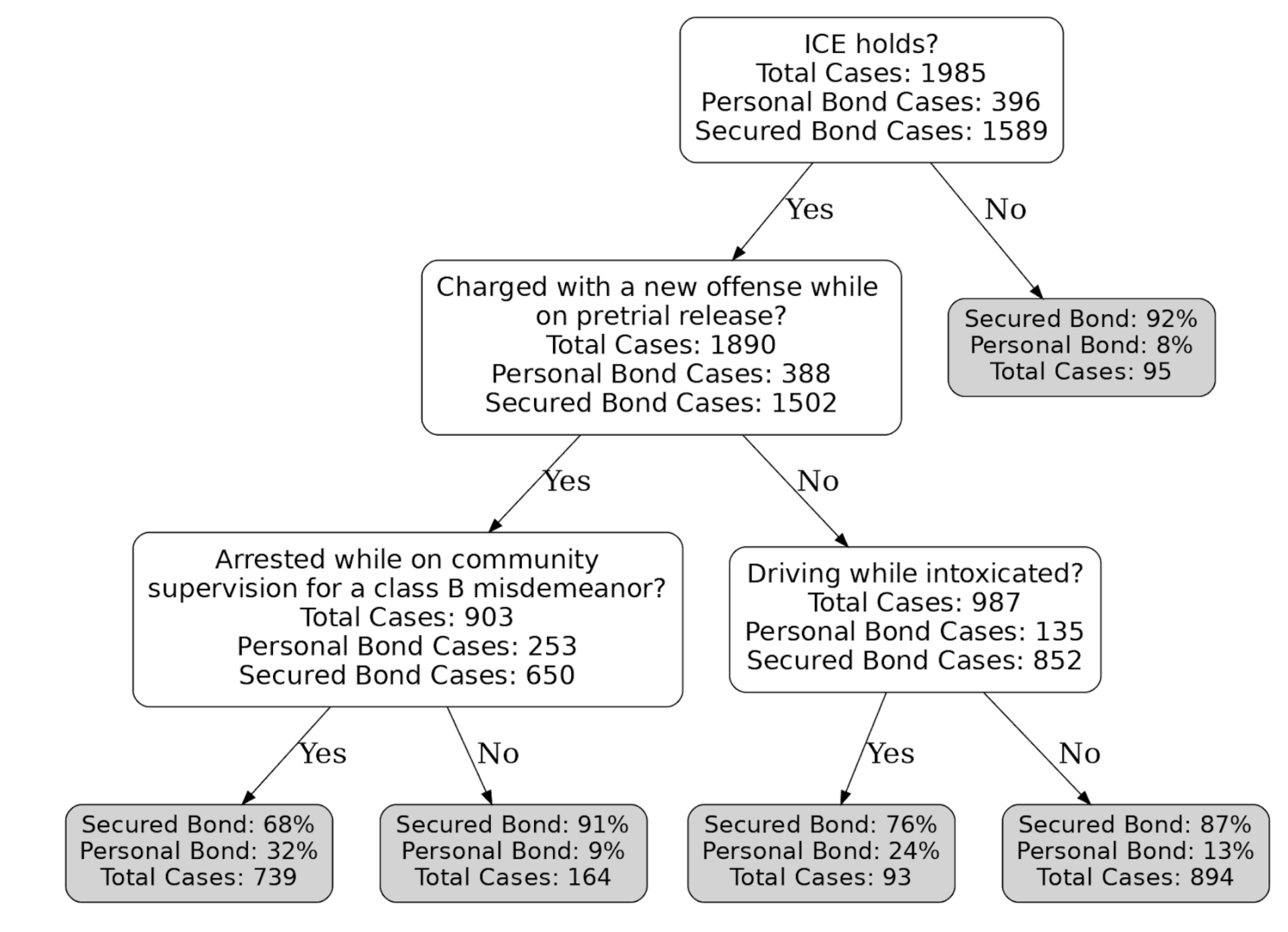}
    \caption{Sample decision tree from the Rashomon set for Judge~17.}
    \label{fig:Judge17Tree3}
\end{figure}

We examine some trees drawn from the Rashomon sets generated by TreeFARMS. Figure \ref{fig:Judge17Tree3} shows an example tree for Judge 17. More examples from the Rashomon set of Judge 17 can be found in Appendix \ref{app:Judge17}. 
One notable pattern that we observe for these judges is that they seem to grant personal bonds to individuals with ICE holds. After reviewing the PC forms, we realized that when a higher authority places a hold, the judge typically anticipates that this authority will handle the defendant’s outcome.
As a result, the judge often releases the defendant on a personal bond. We show trees from other judges later in this manuscript and in Appendix \ref{app:Judge1}.

\begin{figure}[H]
    \centering
    \includegraphics[width=\columnwidth]{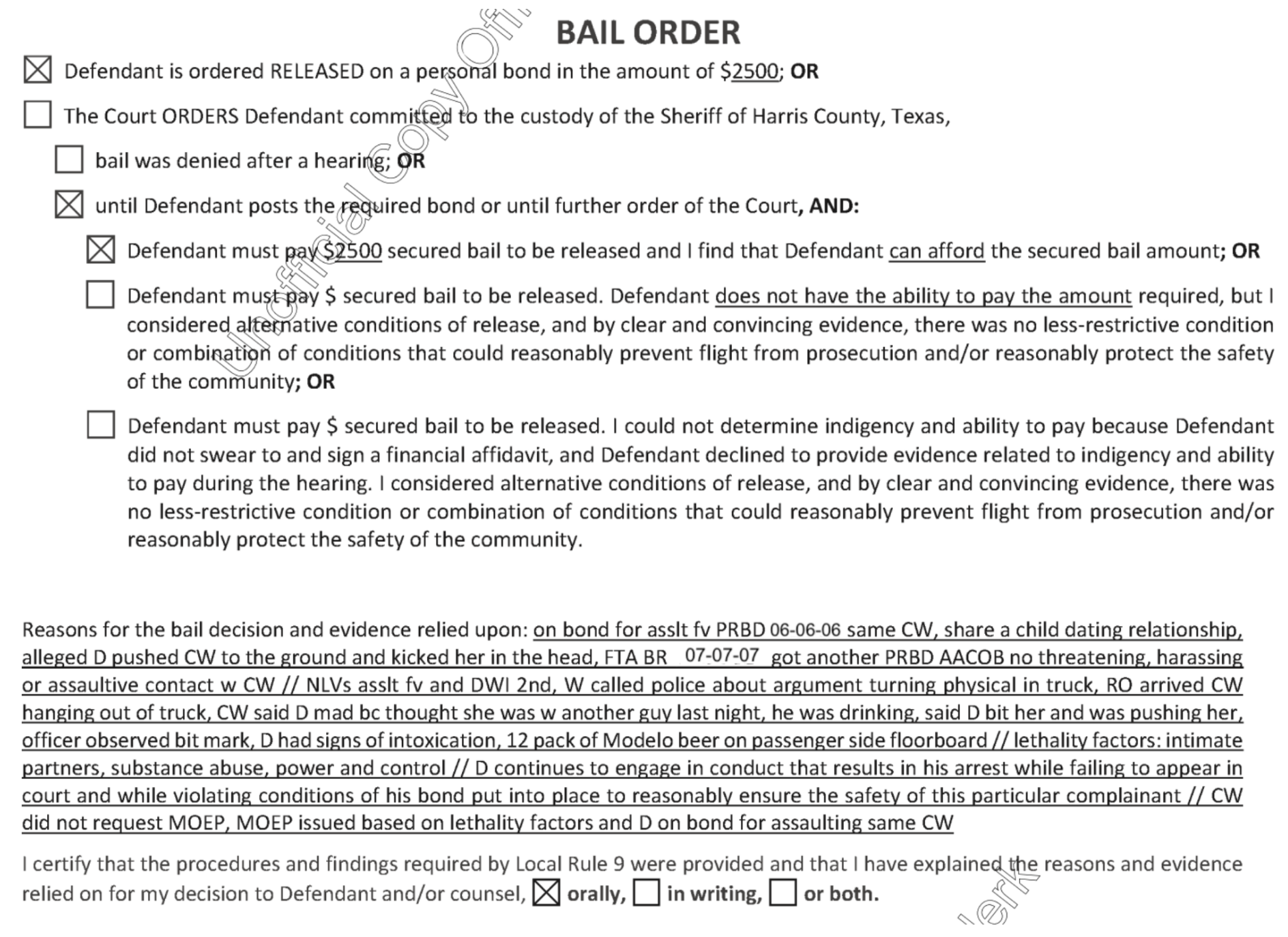}
    \caption{Sample PC form for a case where there seems to be a discrepancy in marking the correct bond type.
    \label{fig:PC1}}
\end{figure}

\begin{figure}[H]
    \centering
    \includegraphics[width=\columnwidth]{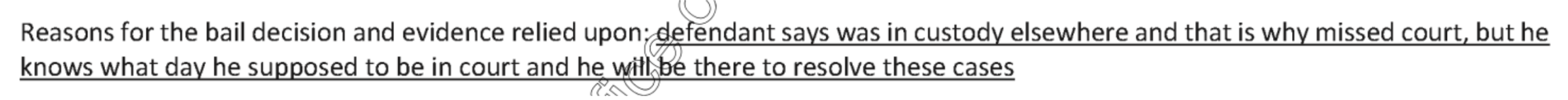}
    \caption{Sample PC form for case where judge had information outside of our dataset influencing their decision.
    \label{fig:PC2}}
\end{figure}
\begin{figure}[H]
    \centering
    \includegraphics[width=\columnwidth]{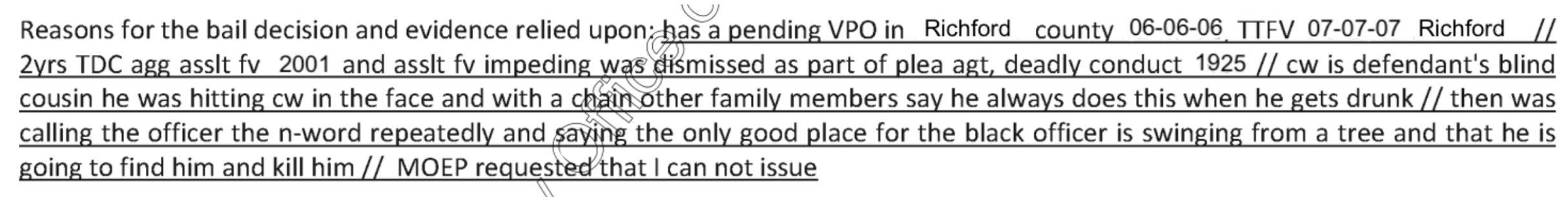}
    \caption{Sample PC form for a case where judge had information about pending charges outside of our dataset, meaning we had incorrect features in our dataset.
    \label{fig:PC3}}
\end{figure}
\begin{figure}[H]
    \centering
    \begin{minipage}{\columnwidth}
        \centering
        \includegraphics[width=\columnwidth]{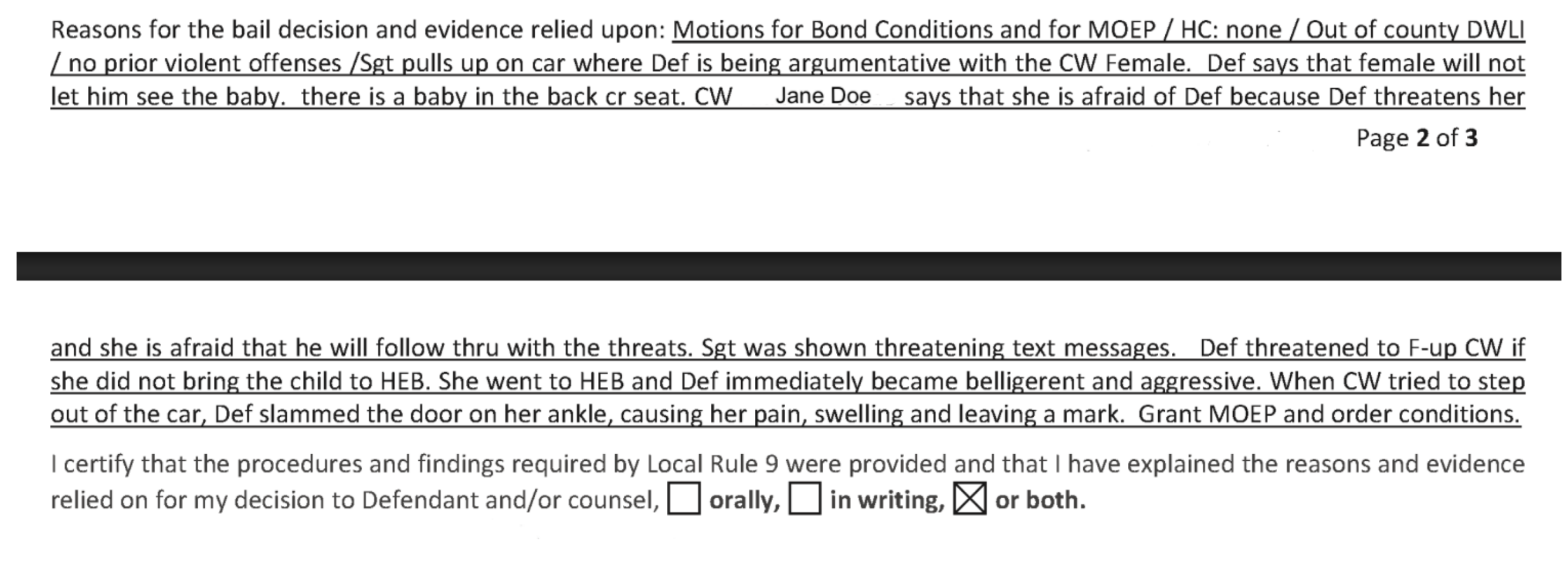}
    \end{minipage}
\rule{\linewidth}{0.5pt}
    \begin{minipage}{\columnwidth}
        \centering
        \includegraphics[width=\columnwidth]{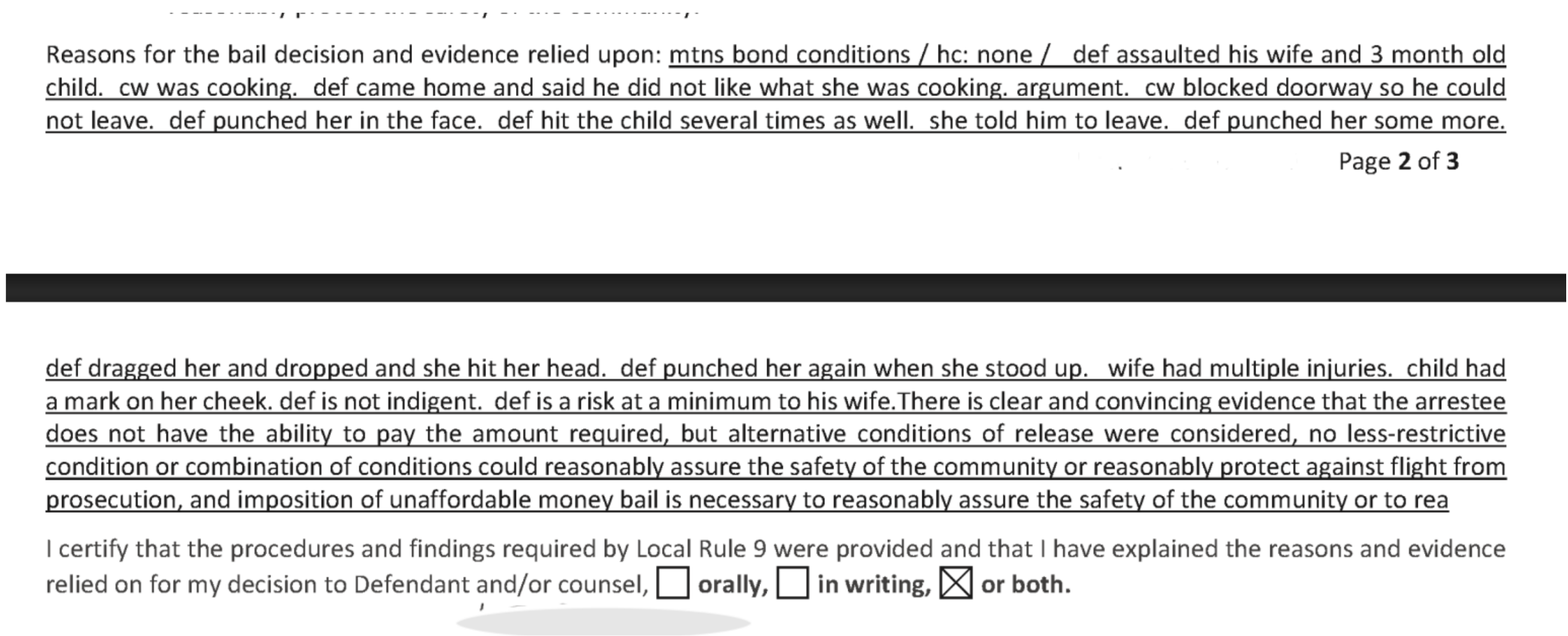}
    \end{minipage}

    \caption{Sample PC forms for cases in which crimes share similar features but result in different outcomes due to differences in crime intensity, a feature we do not have.}
    \label{fig:PC4AND5}
\end{figure}

Next, we used our Rashomon sets to generate subsets of each judge's cases that could not be explained (i.e., predicted correctly) by any model in the Rashomon set. These cases, about 18\% of all cases, are ``unexplainable.'' We investigated a sample of these cases (100) manually and found that all of them were unexplainable due to noise.

We read the judges' justifications and discovered that such cases fell into three main categories: 
(a) Dataset errors (form errors);
(b) Missing information about other crimes (outside of Harris County) that are not in our database;
(c) Details about the case that are not in the database but are available to the judge.

Figure \ref{fig:PC1} shows a form error. Our dataset indicates that a personal bond was granted; however, the judge’s comments suggested against granting a personal bond. When checking the form, we discovered that multiple boxes had been checked for the type of bail order, which must be an error since one cannot receive both a personal and a secured bond in the same hearing. 
The second category includes cases where the judge has information outside of Harris County, which is logistically difficult for us to obtain. Figure \ref{fig:PC2} shows an example where the judge knew that the defendant had been in custody in another jurisdiction, which influenced their decision to grant a personal bond; this information was not available to us in the court data.  
The last category includes cases where the judge makes decisions using details not in our structured data (usually information presented during the hearing but not previously available). Figure \ref{fig:PC4AND5} shows a pair of such cases, where both crimes are Class A misdemeanors of family assault; however, the judge states that they granted a personal bond in the first case and not in the second due to the intensity of the crimes.

All three categories lead to flaws in the data that will not help us understand judges. Ideally, we use data for building models that is identical to the data the judges use. Form errors simply provide incorrect information. Missing information from other counties results in incorrect features in our dataset. 
E.g., if someone commits a crime in a different county, our criminal history features would not record it, creating an error in our data.

The final category is case-specific missing information, such as the severity of the crime, which is not recorded as part of the penal code and crime type (e.g., ``family assault''); this could be remedied if it were consistently mentioned, for instance. 
Including these three classes of cases would require our models to predict decisions using an incomplete representation of the judge's information. This may cause models to attribute the effects of these unobserved factors to the available features, leading to biased estimates of feature importance and poorer generalization. 
Excluding these noisy cases allows models to be trained on observations where the recorded features mainly reflect the available information during a judicial decision and avoids diluting results with noise.  
\begin{figure}[h]
    \centering
    \includegraphics[width=\linewidth]{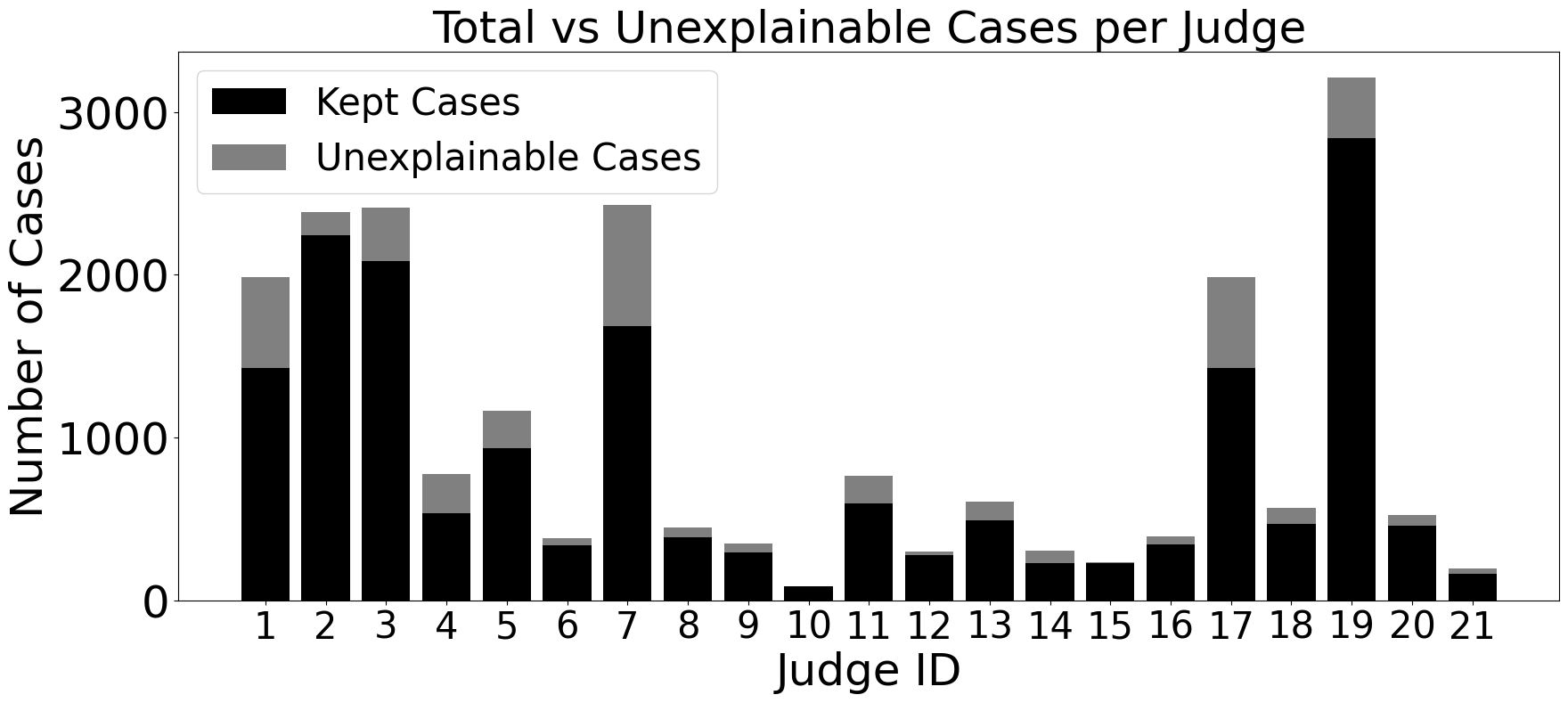}
    \caption{Per-judge distribution of unexplainable vs kept cases.}
    \label{fig:OmittedCases}
\end{figure}

The number of unexplainable cases compared to total cases is shown in Figure \ref{fig:OmittedCases}, and the class distribution before and after filtering is illustrated in Figure \ref{fig:filteringdist} in Appendix \ref{app:Filtered}. 
In what follows, we will remove all the unexplainable cases and only use the remaining cases.
\subsection{Can each judge be described by an algorithm?}
\begin{figure}[htbp]
    \centering
    \includegraphics[width=\linewidth]{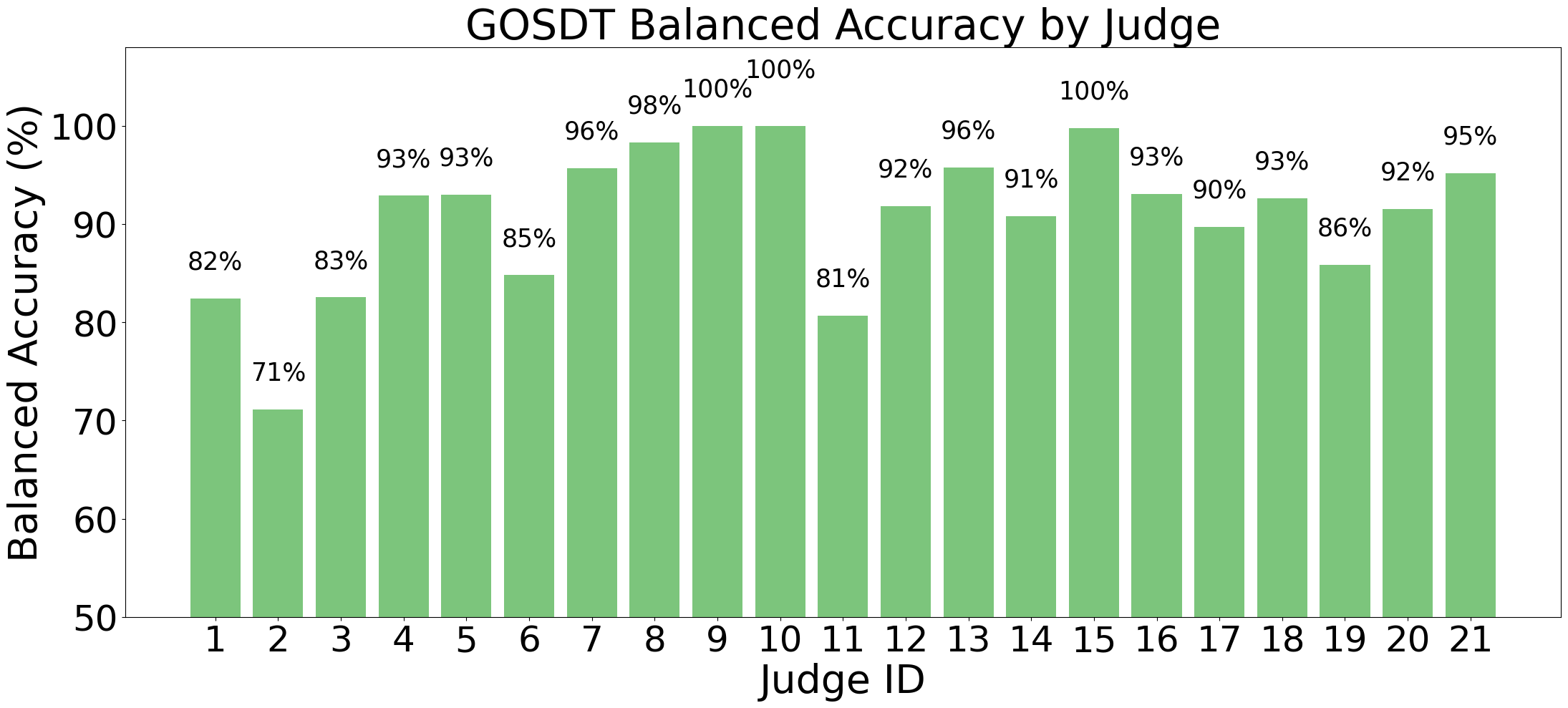}
    \caption{Balanced accuracy for the GOSDT models trained for each judge.}
    \label{fig:GOSDTAcc}
\end{figure}
If a judge follows a systematic method -- an algorithm -- to determine whether to grant a personal bond, we should be able to figure out what that algorithm is. Examining this question allows us to evaluate judicial decision making, including whether the rules judges follow in practice are desirable. Again, we are \textit{not} arguing for replacing judges with algorithms, but rather seeking to identify when and whether judges already behave like algorithms. 
\begin{figure}[htbp]
    \centering
    \includegraphics[width=\linewidth]{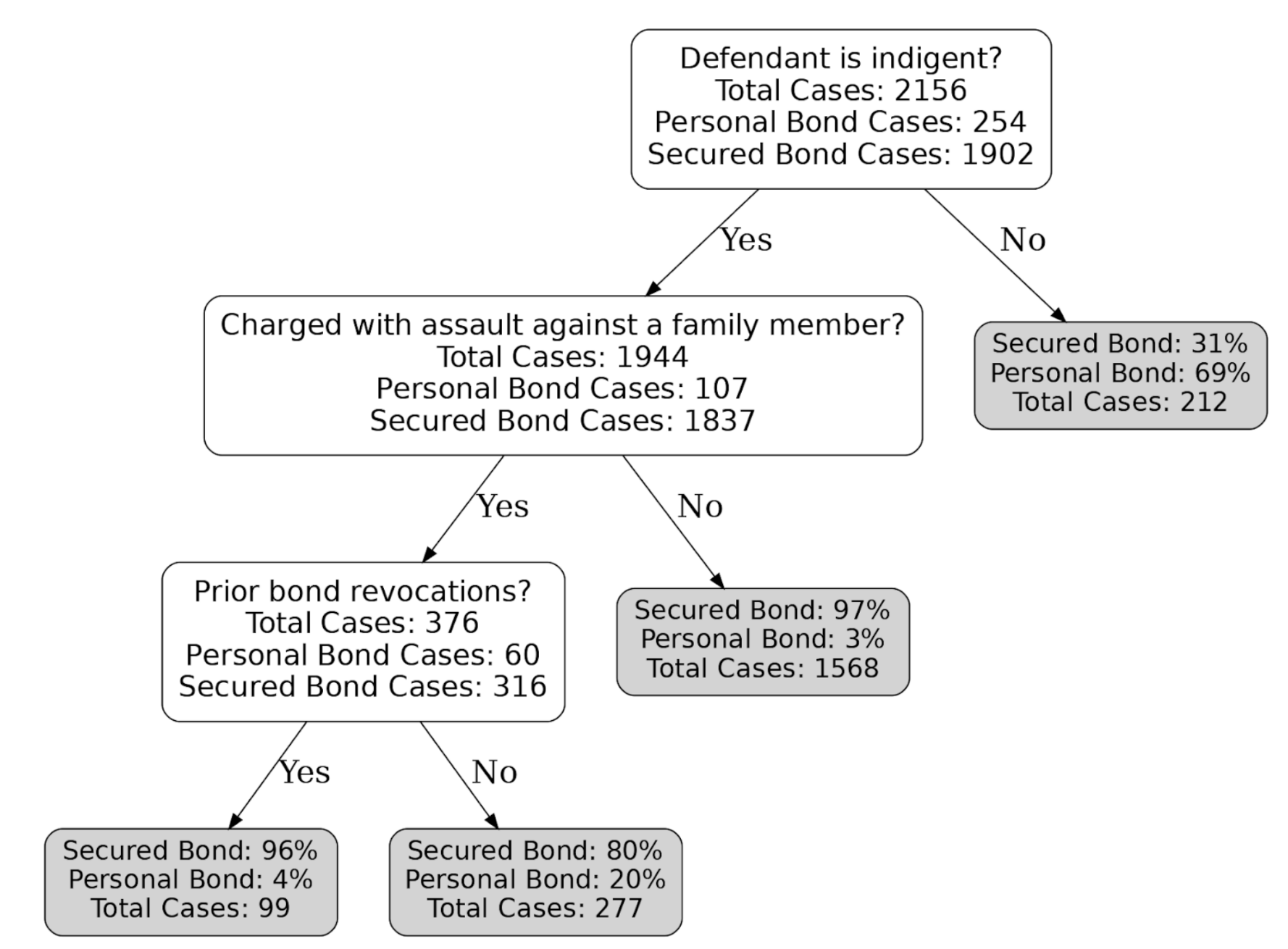}
    \caption{Decision tree for Judge~1 trained using GOSDT.}
    \label{fig:Judge1DT}
\end{figure}

To answer this question,  we trained representative models using GOSDT \citep{LinEtAl20} on the filtered subset of cases per judge, producing globally optimal sparse decision trees. As discussed, decision trees are a powerful, nonparametric function class whose models are understandable. Each magistrate's model was fitted with a tree depth budget of 5, regularization set to 0.001 (a small value, to allow for higher accuracy), and class balancing enabled. This yielded concise, interpretable rules that mimic each judge’s decision style. We report the balanced accuracy in Figure \ref{fig:GOSDTAcc} and show visualizations of some of the resulting decision trees with per-node class counts to interpret the logic of each model.

We discovered that most of the judges could indeed be explained reasonably well by simple algorithms. Figure \ref{fig:Judge4DT} in Appendix \ref{app:gosdt} shows the model for Judge 4 with an accuracy of 91.79\% and a balanced accuracy of 92.88\%. That is, about 92\% of Judge 4’s cases agree with this decision tree, and it is accurate in explaining decisions for both personal bonds and secured bonds. For 17 of the 21 judges, the models capture at least 85\% of the judge’s decisions. Another example of a good model is presented in Figure \ref{fig:Judge1DT} for Judge 1, with an accuracy of 84.37\% and a balanced accuracy of 82.44\%.
Figure \ref{fig:Judge16DT} in Appendix \ref{app:gosdt} shows a model that explains Judge 16 well, with an accuracy of 92.46\% and a balanced accuracy of 93.06\%.

There are several judges whose decision making we cannot explain as well. Appendix \ref{app:BadTree} shows an example of a  tree for Judge 2 that has clearly been overfit (with fewer than 20 cases per leaf) and yet has an accuracy of 77.48\% and a balanced accuracy of 75.41\%. Even after overfitting to the data, 23\% of the cases have no simple rule characterizing them. 

For judges that can be explained, we next examine which factors they seem to consider when making bail decisions. 

\subsection{Do judges consider the same variables?}

In this section, we will show that judges generally consider similar types of variables but differ in the importance of specific variables.

For each judge, we used a bootstrapped feature selection algorithm. For each bootstrap, we balanced the data by oversampling the minority class. Then, we used the Threshold Guessing Binarizer (which uses a boosted tree classifier) described in \citet{mctavish2022fast}, using 50 estimators and a max depth of 3, to transform the variables into binary indicators. Features that did not appear from the Binarizer are unimportant to boosted trees and were removed. Then, we used a boosted tree classifier again on the remaining features. We used this classifier to obtain variable importance scores for each original variable $x_j$, whose importance is denoted by $I_j$. We assigned $I_j=0$ for all unselected features.

\begin{figure}[htbp]
    \centering
    \includegraphics[width=\linewidth]{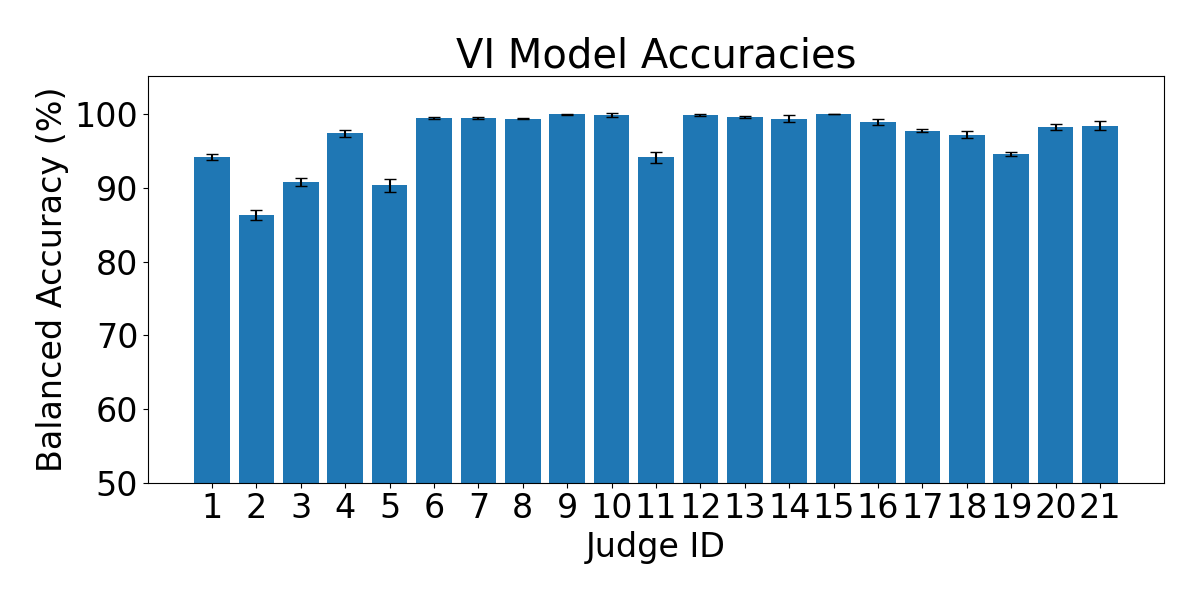}
    \caption{Model accuracy, built from top 50 features, averaged over bootstraps, for each judge. Standard deviations are also shown.}
    \label{fig:VIAcc}
\end{figure}

Across bootstrap resamples, the selection frequency of feature $j$ is
$
F_j = \mathbb{P}(I_j > 0),
$
the proportion of bootstraps in which feature $j$ is used.
The final feature score is
$
S_j = I_j \cdot F_j 
$, and we select the top 50 features
by $S_j$ in descending order.
From these 50 features, we computed the following average variable importance metrics over 100 bootstraps: permutation importance, conditional model reliance (CMR), and leave one covariate out (LOCO). We also computed the accuracy of a model that uses these top 50 features, and we repeated this for each bootstrap and averaged the result over bootstraps. The result is in Figure \ref{fig:VIAcc}.

As we have reduced the noise by filtering out unexplainable cases, we intentionally overfitted the models to the data to ensure that the variables are meaningfully important to the data we have. We see in Figure \ref{fig:VIAcc} that the average accuracy for most judges is extremely high. Reviewing the variables among all judges in Table \ref{tab:VIOverall}, we see that age is \textit{extremely} important, being important for every single judge. Meanwhile, most judges use warrant count and  ``rule'' variables, which we define as flags for specific crimes that are algorithmic in nature for bond trials, indicating that many judges are using them correctly. Furthermore, we can see that race is only important to some judges, allowing us to uncover potential biases that different judges may have. Furthermore, the collection of variables is not consistent among judges, suggesting that judges generally have quite different values when making pretrial decisions. For example, when we compare the top 10 variables for Judges 4 and 6 in Table \ref{tab:VIJudges}, we see a trend where they share many common variables, such as age, warrant count, and ``rule'' variables, but also have unique variables that each weighs heavily. Judge 6 weighs family assault and criminal history more heavily than Judge 4. In terms of demographics, both judges consider age and citizenship status, but Judge 6 appears to weigh gender and homelessness much more than Judge 4, showing how we can use this framework to see the differences in what different judges value.
\begin{table}[th]
\centering
\scriptsize
\setlength{\tabcolsep}{3pt}
\begin{tabularx}{\linewidth}{Xcccc}
\toprule
\textbf{Feature} & \textbf{Appear.} & \textbf{Mean rank} & \textbf{Prop.} & \textbf{Score} \\
\midrule
Age & 21 & 3.238 & 1.000 & 0.309 \\
Warrant Count & 19 & 5.263 & 0.905 & 0.172 \\
Charged with a new offense while on pretrial release? (rule) & 15 & 4.800 & 0.714 & 0.149 \\
Arrested after a bond forfeiture or revocation? (rule) & 13 & 7.359 & 0.619 & 0.084 \\
Arrested while on supervision for a class B misdemeanor? (rule) & 11 & 8.136 & 0.524 & 0.064 \\
First warrant is a CAPIAS warrant? & 10 & 8.667 & 0.476 & 0.055 \\
African American? & 12 & 11.444 & 0.571 & 0.050 \\
Violation of bond conditions for penal code 25.07 (rule) & 9 & 9.500 & 0.429 & 0.045 \\
Family Assault? & 9 & 10.519 & 0.429 & 0.041 \\
How many past felonies? & 9 & 11.037 & 0.429 & 0.039 \\
\bottomrule
\end{tabularx}
\caption{Variable importance summary across judges. Features are ordered by score, defined as $\text{proportion}/\text{mean\_rank}$, where \textit{mean rank} is the average rank when the feature appears in a judge's model and \textit{proportion} is the fraction of judges for which the feature appears in the top 15.
\label{tab:VIOverall}}
\end{table}
\begin{table*}[htbp]
        \centering
        \resizebox{\textwidth}{!}{
        \begin{tabular}{lrrr|lrrr}
        \multicolumn{4}{c}{\textbf{Judge 4}} & \multicolumn{4}{c}{\textbf{Judge 6}} \\
        \toprule
        \textbf{Feature} & \textbf{Rank} & \textbf{CMR}  & \textbf{LOCO} & \textbf{Feature} & \textbf{Rank} & \textbf{CMR}  & \textbf{LOCO} \\
        \midrule
        Age & 1 & 1 & 1.0 & Age & 2.0 & 2 & 1.0 \\
        Arrested after a bond forfeiture or revocation? (rule) & 4 & 3 & 3.0 & Female? & 1.0 & 1 & 10.0 \\
        Criminal Trespassing? & 3 & 4 & 4.0 & Harrasement with previous conviction & 4.0 & 4 & 4.0 \\
        Assault with Bodily Injury? & 5 & 5 & 2.0 & Family Assault? & 7.0 & 5 & 12.5 \\
        ICE hold & 2 & 2 & 8.5 & Unknown Citizenship & 13.0 & 14 & 3.0 \\
        Unknown Citizenship & 6 & 6 & 5.0 & Charged with a new offense while on pretrial release? (rule) & 17.0 & 10 & 5.0 \\
        Warrant Count & 7 & 7 & 7.0 & Warrant Count & 15.0 & 19 & 2.0 \\
        Charged with a new offense while on pretrial release? (rule) & 12 & 8 & 10.0 & Penal Code 28.03 & 11.0 & 9 & 17.5 \\
        Violation of bond conditions for penal code 25.07 (rule) & 10 & 12 & 15.0 & Homeless? & 3.0 & 3 & 33.0 \\
        First warrant is a CAPIAS warrant?  & 14 & 13 & 11.0 & How many past felonies & 9.0 & 17 & 14.0 \\
        \bottomrule
        \end{tabular}}
        \caption{Variable importance ranks for Judge 4 (left) and Judge 16 (right) for the top 10 variables. The variables are ordered by mean ranking over the three importance metrics for each judge, with rank referring to permutation importance. }
        \label{tab:VIJudges}
\end{table*}
\subsection{Could all judges be using similar algorithms? Are judges consistent with themselves and each other?}
In order to ensure fairness in our justice system, we want judges to make decisions consistently. That is, presented with the same information, we want them to make the same decision. We also want different judges to make the same decisions when presented with similar information; i.e., we want all judges to follow the same algorithms. In this section, we hope to analyze the consistency of a judge’s own decisions and those of others, as well as provide a way to give feedback to current judges and judges in the future. To do this, we performed three different sets of experiments.
\subsubsection{Are judges consistent with each other? Pairwise loss distribution analysis.}
We carried out a cross-judge evaluation to examine consistency in decision rules across judges. For each judge, we took the full Rashomon set of near-optimal decision trees and used them as source models. Every model was evaluated not only on the judge’s own cases but also on the cases belonging to every other judge. To make this feasible, we standardized the input space by reconstructing the engineered features for all judges and aligning them by case ID. Any features expected by a source model but missing in a target judge’s dataset were added as zero columns, ensuring comparability across evaluations. Losses were computed separately for the source judge’s data and for each target judge’s data. By repeating this for all models in every Rashomon set, we obtained a full distribution of losses rather than a single point estimate. These distributions were visualized with violin plots.

\begin{figure}[ht]
    \centering
    \includegraphics[width=\linewidth]{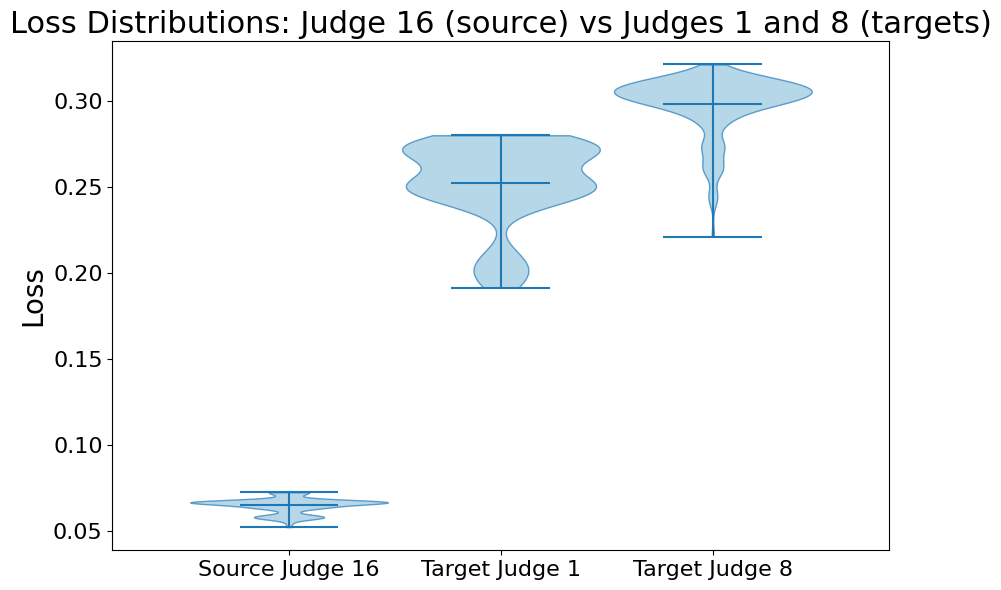}
    \caption{Rashomon model loss distributions on Judge~16 source cases. Compares the source-judge model (S16) with target-judge models (T1 and T8).}
    \label{fig:Loss_S16}
    
\end{figure}

We found that most of the judges are not consistent with one another. For example, Figure \ref{fig:Loss_S16} shows the violin plot of the loss distribution for all the models within the Rashomon set of Judge 16 when evaluated on their own cases (source cases) versus those of Judges 1 and 8 (target judges). We see that the models perform consistently well on the source data, with the error rates clustering tightly around 6\%, which indicates coherent decision logic. However, when the same models are applied to the target judges, the losses see a sharp rise, with larger ranges and more errors for the target judges. Although the model found for Judge 16 captures multiple near-optimal explanations for their own decisions, these explanations do not apply to the decisions made by Judges 1 and 8. This contrast suggests that both judges behave differently from Judge 16 in their decision-making processes.

However, in some judge pairs, the difference between the source and target distributions is less pronounced, as shown in Figure \ref{fig:Loss_other_pairs} (left) in Appendix \ref{app:loss}. Here, the Rashomon models of Judge 2 maintain comparable performance when evaluated on Judge 3, with overlapping distributions and similar median error rates of 0.24 to 0.27. While the variance increases slightly when predicting on the target judge's cases, the overall shape of the distributions suggests partial alignment in decision behavior between the two judges. This implies that both judges may rely on the same features and follow similar reasoning patterns when deciding cases. Figure \ref{fig:Loss_other_pairs} (left) also shows that Judge 2 may be internally inconsistent since their loss distribution seems to be quite spread-out. 
We also found several examples with partial overlap. The violin plots in Figure \ref{fig:Loss_other_pairs} (right) show that while the Rashomon models of Judge 6 perform reasonably well on their own decisions, their performance varies widely when applied to Judge 15. The source distribution is reasonably compact, with a median error around 0.20. In contrast, the target distribution is widest near both the lowest and highest ends of the loss scale, suggesting a bimodal pattern of generalization. Some models from the source set transfer well, achieving low errors on the target judge, while others perform poorly, reaching error rates above 0.50. This pattern could imply that the two judges share overlapping logic on certain subsets of cases but may diverge on others. 

\subsubsection{When do judges disagree? Rule mining analysis.}
We then conducted rule mining to extract interpretable conditions under which specific pairs of judges’ decision rules diverged. For each ordered judge pair, we used majority-vote predictions from both Rashomon sets on the source judge’s cases to define a binary disagreement variable, which would be true if the labels did not match and false otherwise.

In rule mining, \textbf{support} measures how frequently an itemset appears in the dataset, \textbf{confidence} indicates how often the rule’s consequent occurs given its antecedent, and \textbf{lift} quantifies how much more likely the consequent is to occur with the antecedent than it would be by chance.
\begin{table}[ht]
\centering
\scriptsize
\setlength{\tabcolsep}{4pt}
\renewcommand{\arraystretch}{1.15}

\begin{tabularx}{\columnwidth}{p{0.9cm} p{0.9cm} X c c c p{0.9cm}}
\toprule
\textbf{Judge A} & \textbf{Judge B} & \textbf{Disagreement Rules} &
\textbf{Supp.} & \textbf{Conf.} & \textbf{Lift} & \textbf{Cases} \\
\midrule
Judge 4  & Judge 7  & Age $\leq 24$ AND assault on a family member AND $\leq 1$ prior warrant & 0.193 & 0.935 & 1.70 & 305/326 \\
Judge 1  & Judge 17 & Assault on a family member AND $\leq 1$ prior felony AND new offense on supervision & 0.229 & 0.852 & 2.09 & 420/493 \\
Judge 9  & Judge 14 & $\leq 1$ past crime & 0.623 & 0.750 & 1.43 & 138/184 \\
Judge 2  & Judge 18 & Indigent AND first arrest due to non-appearance AND $\leq 1$ prior warrant & 0.195 & 0.769 & 1.81 & 337/438 \\
Judge 7  & Judge 9  & first warrant issued for crime commitment & 0.356 & 0.757 & 1.15 & 69/105 \\
\bottomrule
\end{tabularx}

\caption{Five sample rules identified by FP-Growth that characterize situations where pairs of judges disagree.}
\label{tab:rules}
\end{table}
The antecedent space is defined as the union of the engineered binary features used by both judges. We mined candidate rules of length one to three under minimum support (0.02 -- meaning 2\% of the data must obey the left side of the rule) and minimum confidence (0.60 -- meaning at least 60\% of the right side of the rule obeys it) thresholds, with the top 50 rules per length. 

We used FP-Growth \citep{han2000} to generate our set of rules for each judge pair. Each rule was then used on the underlying cases to compute canonical counts of covered cases, disagreements, and agreements, from which support, confidence, and lift were recomputed relative to the global disagreement rate. This procedure, applied in both directions for all judge pairs, yielded a set of concise, understandable rules highlighting the specific feature combinations most predictive of inter-judge disagreement.

We find that most judges with distinct loss distributions exhibit strong, interpretable rules that explain their disagreements. For example, the rule ``age $\leq$ 24 $\rightarrow$ Judges 9 and 4 disagree'' is highly predictive (support = 0.18, confidence = 0.94, lift = 1.67), indicating that disagreements between Judges 9 and 4 occur in nearly all cases involving defendants aged 24 or younger. Similarly, the rule ``charged with family assault AND charged with new offense on pretrial release $\rightarrow$ Judges 21 and 17 disagree'' identifies a substantial subset of cases: 24\% satisfy both conditions, and the judges disagree in 85\% of them. These patterns suggest that outcomes in such cases are driven more by individualized judicial discretion rather than by standardized risk cues. These rules can identify specific case-qualities where judges most often disagree, providing an avenue to understand when judges disagree and potentially reduce inconsistencies in their decision-making by aligning judge's decisions in these scenarios.

Additional examples are in Table \ref{tab:rules}.

\subsubsection{Are judges more
consistent with their own decisions than with others? Triplet analysis.}
\begin{figure}[H]
    \centering
    \includegraphics[width=\columnwidth]{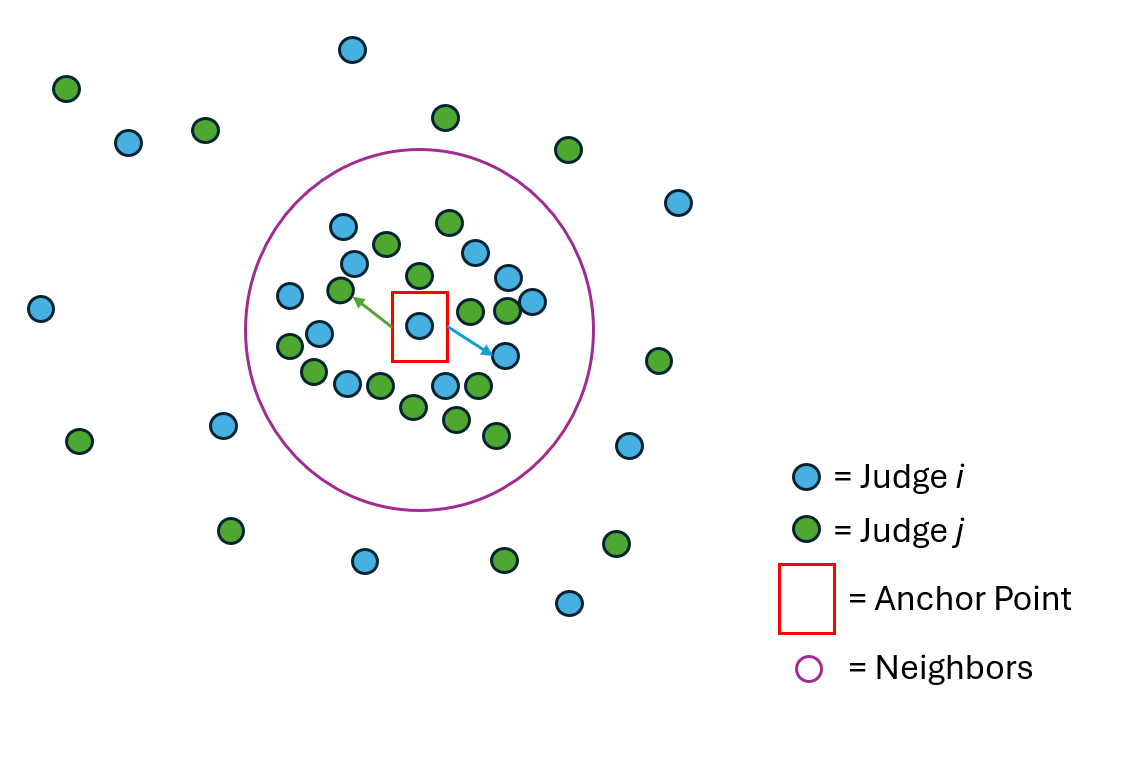}
    \caption{Triplet Example. This is an example of a triplet, where the blue points are from Judge $i$ and the green points are from Judge $j$. Given an anchor point from Judge i, a triplet is then made from choosing a random nearest neighbor from Judge $i$ and Judge $j$. In this example, any point within the purple circle is available to be chosen as a random nearest neighbor for the triplet. 
    \label{fig:triplet}}
\end{figure}
 Even if two different judges follow different algorithms, it is still important to see whether they make similar decisions. To do this, we performed a triplet analysis that compares two judges. For each pair of judges, there is an anchor judge and a target judge. A triplet is defined by an anchor case from the anchor judge, a nearest neighbor case from the anchor judge, and a case from the target judge that is similar to the anchor case. We define similar cases to  be cases with a low normalized weighted Euclidean distance, where highly important variables like offense type, rules, and age are weighted more heavily than other variables. Figure \ref{fig:triplet} shows a visualization of how these triplets are formed. In this triplet analysis, we randomly sampled $\sim$1000 triplets for each pair of anchor and target judges. Analyzing these triplets, we consider the following questions: \textit{When two judges disagree, does each judge predict themselves better than the other?}, \textit{When two cases presented to the same judge are similar, do they make the same decision?}, and \textit{Does the choice of judge affect the decision of a case?} Let us answer these. We use the same triplets in our entire analysis so that we can directly compare our metrics.  
\paragraph{When two judges disagree, does each judge predict themselves better than the other?
}

\begin{figure}[ht]
    \centering
    \includegraphics[width=\columnwidth]{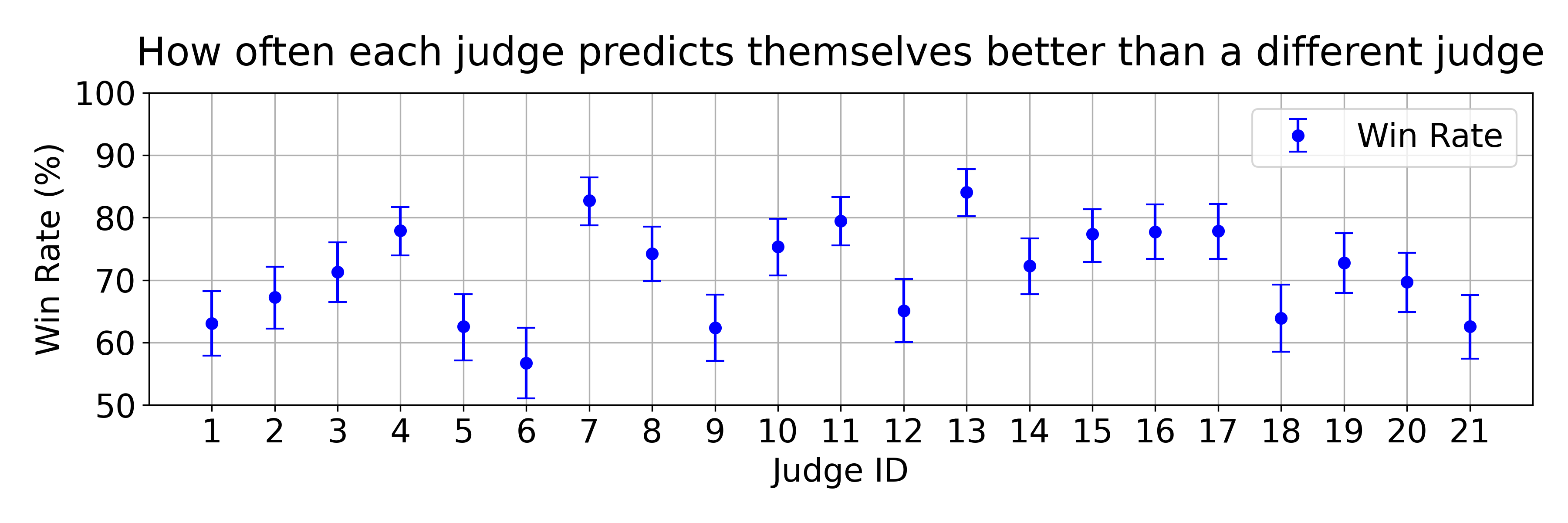}
    \caption{Average win rates of  judges. All win rates exceed 0.5, indicating that judges predict their own outcomes better than those of other judges in the majority of cases.}
    \label{fig:winrate}
\end{figure}

To evaluate this, we calculate the percentage of triplets where Judge $i$ makes the same decision as the anchor point, meaning Judge $i$ predicts themselves correctly, while Judge $j$ disagrees. We define this metric as the win rate. If a judge predicts themselves better than other judges do, they have a high win rate. We see in Figure \ref{fig:winrate} that most judges have a win rate between 60\% and 80\%, meaning that each judge is a better predictor of themselves than other judges most of the time. While this is good, we also see that there can be improvement, as the win rates cap out at around 85\%.

\paragraph{When two cases presented to the same judge are similar, do they make the same decision?}
\begin{figure}[ht]
    \centering
    \includegraphics[width=\columnwidth]{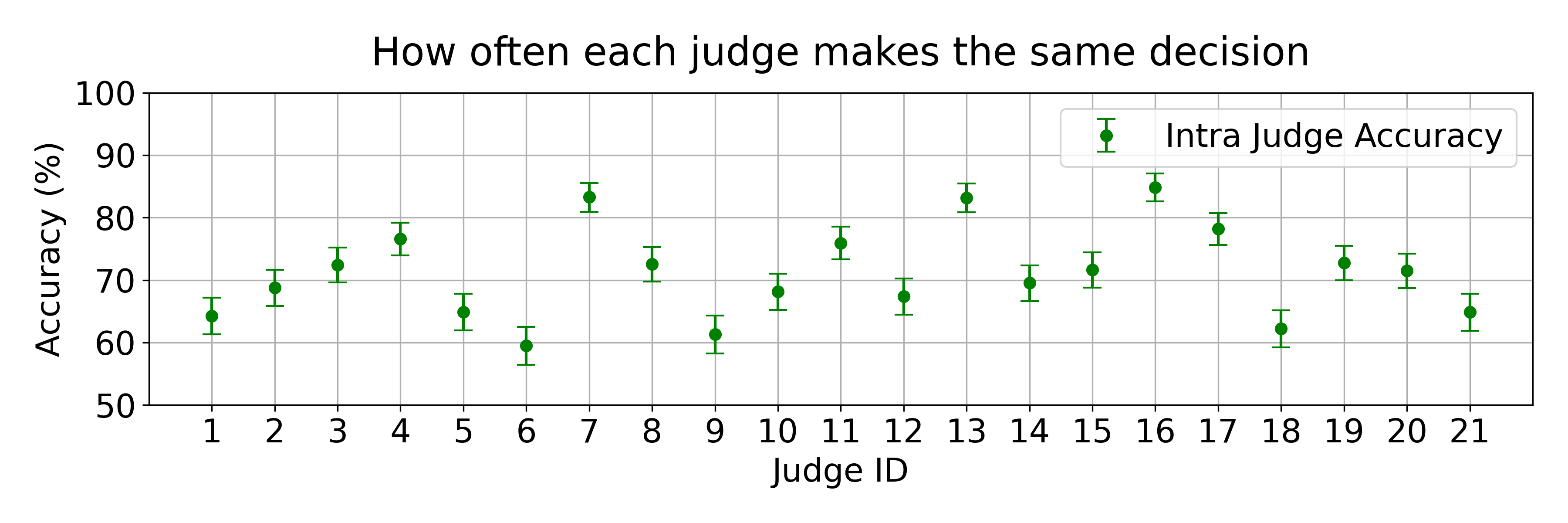}
    \caption{Judge's consistency in decision-making. The average percentage of triplets where a judge predicts the same outcome as the anchor case. All judges are above 50\%, indicating performance better than random guessing, with most around 60–70\% and some near 80\%. 
    \label{fig:intra}}
\end{figure}

For a fair justice system, we expect a judge to always make the same decision when given the same case. To evaluate whether this happens or not, we find the percentage of triplets (using just the anchor and the other point from the same judge) where the anchor judge makes the same decision. As we see in Figure \ref{fig:intra}, most judges have a percentage between 60\% and 80\%, indicating that while they are generally more consistent than random, there is still a large gap between judges always being consistent with themselves. This is an important evaluation to  ensure that judges are maintaining consistency.

\begin{figure}[th]
  \centering
    \includegraphics[width=\columnwidth]{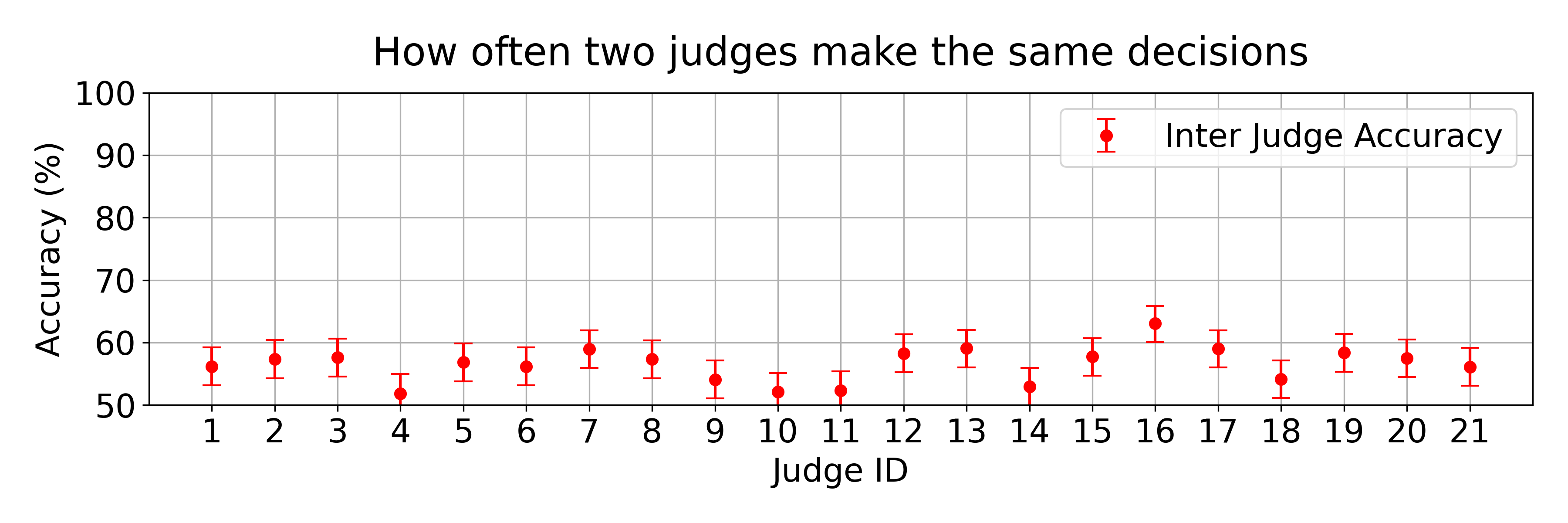}
    \caption{Consistency between judges. The average percentage of triplets where a judge predicts the same outcome as the second judge in the triplet. We see that all judges are slightly better than random guessing. This shows that agreement between judges is somewhat random.
    \label{fig:inter}}
\end{figure}

\paragraph{Does the choice of judge affect the decision of the same case?}
While we cannot test giving two different judges the same case, we can use our triplets to approximate this. To evaluate whether the choice of judge matters, we find the percentage of triplets (using just the anchor from Judge $i$ and the point from Judge $j$) where Judge $i$ makes the same decision as Judge $j$. A higher percentage means the choice of judge does not matter, while a low percentage near 50\% suggests that it does. As we see in Figure \ref{fig:inter}, the percentage is about 50\%, maxing out at $\sim$ 60\%, which is the minimum percentage for both the win rate and how consistent each judge is with themselves. This means that the choice of judge is \textit{very} important in deciding whether a defendant will receive a personal bond or not.

\section{Discussion}
In our justice system, we expect judges to make decisions consistently. Thus, we expect judges to follow \textit{rules}: decision patterns using algorithmic logic. However, some cases require the application of \textit{standards}, i.e., contextual understanding and moral reasoning that computational models cannot replicate. If judges do follow rules, we can potentially improve them through better predictive modeling; improving rules is comparatively inexpensive and quite feasible \citep{ludwig2024}, and would make judicial decisions more accurate \citep{demichele} and transparent. If judges do not follow algorithms, it is important to know why, because arbitrariness in judicial decision-making is undesirable. We have provided and used a framework that evaluates how often judges follow algorithms and how and whether their algorithms differ. We saw that judges often follow algorithms and that different judges follow different algorithms, which produce different outcomes in similar cases. This means judges could potentially be made more consistent and accurate. The framework also includes a variable importance analysis that can uncover whether judges may heavily consider features like race or other demographic characteristics to make decisions. We also found many cases where missing or erroneous data required the use of standards rather than rules. Better data collection could then lead to more consistent, rule-based decisions. Given high quality recidivism data, it is possible to extend the framework by evaluating judge's algorithms to determine whether judges accurately predict long-term outcomes. 

Instead of viewing algorithms as replacements, they should be seen as instruments to inform and enhance human judgment. No one can accurately estimate risks from sufficiently large and complex data in their head, but ensuring human oversight remains central, especially in high-stakes cases that could determine an individual's freedom. By making judges’ internal rules, i.e., algorithms, as well as their reliance on standards, more explicit and accountable, we can move towards a judicial system that is both transparent and self-improving.
\section{Acknowledgements}
We thank Arvindh Manian for his contributions to the early stages of this work.

\bibliography{aaai2026}
\clearpage
\appendix
\setcounter{secnumdepth}{2}
\setcounter{section}{0}

\renewcommand{\thesection}{\Alph{section}}
\renewcommand{\thesubsection}{\thesection.\arabic{subsection}}
\renewcommand{\thesubsubsection}{\thesubsection.\arabic{subsubsection}}
\section*{Appendix}
\section{Further Background and Related Works}
\subsection*{Benefits of rules and standards}
\label{app:RulesStandards}
\subsubsection{Rules}
As discussed in the introduction, there is a broader debate about whether judges, and the judicial system, should rely more on rules versus standards. To provide an example from the misdemeanor bail system in Harris County of a rule, the O'Donnell Consent Decree imposes, and the local rule that the judges follow similarly adopts, a rule that, for non-carve-out misdemeanor offenses, a person is to be promptly released on a general order bond. It is not discretionary; the rule is applied automatically. 

An example of a standard is the Consent Decree’s rule, reflecting the requirements of procedural due process, that a judge must find ``clear and convincing evidence'' that a person poses a risk of flight or to public safety in order to impose a secured bond in a misdemeanor case. That standard supplies a burden of persuasion and notes the relevant considerations, but it does not provide specific directions regarding how a magistrate should make decisions in particular cases. Thus, magistrates retain substantial discretion in individual cases when deciding whether the evidence reaches the clear and convincing level regarding the dangers of flight or public safety. 

One way to think about rules versus standards is that they provide different degrees of discretion for the judge as a decision-maker. Rules and standards exist on a continuum, and a given rule may have elements that retain some flexibility or discretion, for example. However, the role of a judge is far less important the more a bright-line rule is relied upon. A judge does not exercise independent judgment if there is a clearly applicable rule, except to identify the predicate facts making the rule applicable. Rules limit discretion and enforce consistency. Getting rules right in advance, and avoiding the resulting over- and under-inclusiveness, raises real challenges. Standards are appealing where individualized circumstances might vary and be highly relevant, and where equitable discretion, including regarding moral decisionmaking, is highly desirable.

Similar considerations apply to the debate regarding the desirability of algorithms in the judicial system. Let us discuss the reasons why algorithms are useful; the reasons largely track the advantages of rules as compared to standards, but with several more specific issues.

\textbf{Algorithms are consistent:} Two identical cases will be scored the same way. This cannot necessarily be said for two identical cases with different human judges, or even two identical cases presented to the same human judge. And, it may be that an algorithm has advantages that simple judicial rules do not: an algorithm can be more nuanced and take account of a large number of factors and different weights, while traditional legal rules may tend to be more under- and over-inclusive. 

\textbf{Algorithms are transparent (when they are interpretable):} Algorithms for risk scoring should be simple, interpretable, public, functions, whereas judges are black boxes – we do not know their reasoning processes. Algorithms in criminal justice have been harshly criticized, mainly after the coverage of the use of a proprietary (i.e. black box) risk scoring instrument called COMPAS, the Correctional Offender Management Profiling for Alternative Sanctions \citep{AngwinLaMaKi16}. That algorithm was not transparent; if it were, and judges and lawyers knew that it largely depended on a person’s age and criminal history, the debate about its fairness might not have occurred \citep{RudinWaCo2020}. People might have demanded a more useful and nuanced risk assessment if they were to use one \citep{WangHanEtAl2022}. However, most risk assessment tools in criminal justice are transparent, and we should be using only those tools.

\textbf{Algorithms are easier to fix.} While algorithms can be biased, they are much easier to fix than judges, whose rulings may be appealed, but with typically deferential standards of review. Judges are human, as we are all biased in some respects, even when we must provide reasoned explanations for our decisions. Simple formulas do not have this problem.

\textbf{Algorithms are easier to assess for bias.} Human decision-makers may be affected by a range of cognitive biases, and in legal decision making, this is true as in any other area of decision making. Studies have examined the effects of biasing information on legal decision-makers in a range of contexts. For example, \citet{sacks2015} demonstrated that in the U.S., Black and Hispanic defendants were more likely than white defendants to face financial bail requirements and less likely to secure release, highlighting racial disparities in pretrial detention. Algorithms are easier to assess for harmful biases, particularly when their formulas are interpretable. There is now a huge literature on assessing algorithms for fairness, interpretability and accuracy \citep{WangHanEtAl2022} (and creating algorithms that perform well on all of these metrics simultaneously). These evaluations assess algorithms on huge datasets -- such evaluations cannot be performed on individual judges because we cannot possibly know what an individual judge would decide on a huge number of cases -- unless they behave like an algorithm. 

\textbf{Algorithms are more accountable.} As long as the algorithms are transparent and public, there is simply a much greater level of accountability available than there is for individual judges. That is, algorithms \textbf{\textit{allow us to debate}} about whether they are accurate and fair. The same is much more difficult to do for a judge. 

There is evidence that a range of experts, including judges, may act in ways that resemble interpretable algorithms. \citet{dhami2001} analyzes bail decisions in the UK using the lens of fast and frugal trees (FFTs). Findings show that judges’ bail decisions can be accurately modeled using simple decision trees that rely on a small number of cues, suggesting that judicial decision-making is systematically heuristic rather than random or overly complex. Similarly, research in other high-stakes contexts has shown that experts can be well-approximated by compact decision rules. \citet{banks2020} shows how simple decision rules can be trained and applied in high-stakes military contexts. Military officers trained on fast and frugal heuristics reached decisions as accurately as those trained in more complex methods. 

There is a separate question whether simple rules perform better than judges’ existing exercise of discretion. Recent work has focused not only on comparing human and algorithmic decision-making but also on designing interpretable rules that can be applied in real-world practices. \citet{kleinberg2017} show that machine learning models trained on bail data could reduce crime by as much as 24.7\% without raising incarceration rates, or alternatively, reduce jailing by up to 41.9\% without increasing crime, while simultaneously lowering racial disparities. \citet{cofone2021} surveys the integration of AI into judicial processes, distinguishing between judges using AI tools and AI replacing judges. \citet{jung2020} introduce a simple ``select, regress, and round'' procedure for constructing scoring rules that experts can use mentally. Even with just a few features and integer weights, these rules can achieve consistency and transparency, reducing unnecessary detention without raising failure-to-appear rates.
\subsubsection{Standards}
The alternative to relying on algorithms is to rely on judicial discretion. The longstanding rules vs. standards debate can inform these choices.  If there is information that is highly relevant, including morally relevant, that is not readily available to train algorithms, then we would particularly want judges to make more individualized decisions relying on such information. If we do not think that algorithms can, for other reasons, capture factors that are informative in individual cases, then we will similarly prize discretion and individualized justice, rather than more rule-based justice. 

We may prioritize individualized justice for moral and legitimacy reasons as well, viewing human decision making as morally superior and more legitimate, than a more rule-based or algorithmic basis for decision making. Accuracy may not be the only, or the most important, goal in a judicial system. As we noted in the introduction, due process rights entitle people to obtain review from impartial decision-makers in many circumstances when the government acts. That is why ``black box'' algorithms are so troubling, since human decision-makers cannot understand, much less review, their functioning.

Finally, we may view rules as preferable in many or most cases, but not in all cases. We may want a human decision maker to follow a role in many or most cases, but in other cases, we may want the judge to follow a standard, and take into account additional information. We may want that standard to apply in specific types of cases, identified in advance. That said, algorithms can be designed to be complex, far more so than traditional simple legal rules, and to address these more specific situations. Or, we may also want the human decision maker to have the power to override the recommendations of the rule or algorithm when errors in data, or unusual case-specific data, mean that relying on the rule or algorithm would be unreliable. The question that we focus on here is whether judges already follow rule-like approaches, or standards, or some identifiable combination of the two.

\section{Data and Processing}
\label{app:data}
\subsection{Magistration in Harris County, Texas}

In 2019, after several years of federal civil rights litigation in the case of O'Donnell v. Harris County, the County entered a Consent Decree requiring comprehensive reforms of its misdemeanor bail system. The outcomes before and after those reforms have been discussed in some detail previously, including by two of the authors, who serve on a team independently monitoring progress under that Consent Decree \citep{garrett2024}. About 50,000 misdemeanor cases have been filed per year, in recent years, in Harris County. Those cases are handled by the Harris County Criminal Courts at Law (CCCL).  

Under the prior system in Harris County, magistrates did not have real discretion when assessing pretrial detention and bail.  The Texas Constitution promises bail to almost all persons charged with crimes. \cite[][Art. I, 11a-c]{ConstTex}.  A bond is a promise, after arrest, to return to court in a criminal case.  Such a bond can be secured by a money payment made in advance, called bail in Texas, or it can be unsecured and not require any payment up front, which is called a ``personal bond,'' in Texas. \cite[][Art 17.01-2]{CodeTex} Texas law, under Article 15.17 \cite[][Art 15.17]{CodeTex}
of the Texas Code of Criminal Procedure sets out that a magistrate must find probable cause, and then assess whether a person is eligible for bail. \cite[][Art 12.17]{CodeTex}
Texas law has set out factors to be considered when setting bail amounts, including the ``nature of the offense,'' the ``arrestee's ability to make bail,'' the ``future safety of the victim,'' and a state ``public safety report'' with information regarding a person’s criminal history and court appearance \cite[][Art 15.17]{CodeTex}. Texas law had long permitted local courts to adopt bail ``schedules,'' in which local judges made predetermined decisions about setting bonds for specified offenses (although this law has since changed). That is what Harris County had in place before the Consent Decree.  After arrest, if the District Attorney’s Office accepted the charges, bail would be set according to the written schedule adopted by the judges. It had a $\$$500 base amount for misdemeanors, with additional $\$$500 increments added based on the charges and the person’s criminal history.  If a person could pay that amount, they could be released.  In most cases, a person could not.  They received a misdemeanor bail hearing before a magistrate, which would typically last for a minute or two, where magistrates, in the vast majority of cases adhered to the bail schedule amounts \cite[][p.17]{garrett2024}.

After the ODonnell class action was filed, and relief was obtained culminating in a Consent Decree, entered in 2019, magistrates had discretion to make individualized decisions in bail hearings. Most misdemeanor arrestees are promptly released on a non-secured bond after booking; those persons do not receive or need a bail hearing.  However, not all people qualify for this type of prompt administrative release, under the Consent Decree.  Six categories of people arrested for misdemeanors in Harris County, who belong to ``carve-out’’ categories, instead proceed to a bail hearing before a magistrate: persons charged with protective order and bond condition violations; domestic violence cases; repeat driving while intoxicated cases; people charged with a new offense while on pretrial release; people arrested after a bond forfeiture or revocation; and persons arrested while on community supervision \cite[][p.21]{garrett2024}.

Before this bail hearing, Pretrial Services conducts interviews with arrestees, and provides magistrates with an advance report regarding the person’s financial and other circumstances, along with recommendations whether a person should be released or not on a personal bond.  Some of those people are granted a personal bond by the magistrate without a need for a bail hearing, in a process termed ``early presentment'' \cite[][p.21]{garrett2024}. The others proceed to a hearing, held at the Harris County Joint Process Center, during magistration dockets that run all day and night. All misdemeanor arrestees have access to a public defender, who represents them at the bail hearing. 

The Consent Decree and the judge’s rule that accompanies it, require that magistrates use their discretion to decide whether there is ``clear and convincing'' evidence that a person poses a risk of flight or a threat to public safety, in order to order a secured bond in a misdemeanor case.  They must make findings regarding their reasons for imposing any secured bond in a case, either orally on the record or in writing. Those reasoned findings are not required if they impose a personal bond. Their rulings may be reviewed, on the next business day, by a CCCL judge, who may revisit the bond conditions imposed.

The Consent Decree also provides that the public, and the Monitor, be provided access to certain data concerning misdemeanor pretrial release and detention decisions, including arrestee demographic information, which magistrates and judges are assigned to cases, decisions made at bail hearings, and outcomes in cases, from 2009 to the present.

\subsection{Data Description}
\label{app:datadesc}

\begin{table*}[ht]
\centering
\begin{tabular}{p{4cm} p{10cm}}
\toprule
\textbf{Features} & \textbf{Description} \\
\midrule

\textit{past\_mrp, past\_mrp\_f} &
Indicator of whether defendant had past missed required probations/appearances \\

\textit{past\_revokes, past\_revokes\_f} &
Count of prior bond revokals \\

\textit{repeat\_count, repeat\_count\_f} &
Number of times the same penal code appeared in prior cases \\

\textit{num\_pending\_crimes, num\_pending\_crimes\_f} &
Count of cases filed before judge time that were still pending \\

\textit{disp\_dadj, disp\_dadj\_f} &
Count of prior deferred adjudication dispositions \\

\textit{disp\_dadj\_delta, disp\_dadj\_delta\_f} &
Count of deferred adjucation dispositions within a recent time window \\

\textit{num\_past\_crimes, num\_past\_crimes\_f} &
Total number of prior criminal cases \\

\textit{num\_past\_crimes\_delta, num\_past\_crimes\_delta\_f} &
Number of past crimes within a recent time window \\

\textit{num\_past\_for\_caseid} &
Number of prior hearings for the same case \\

\textit{has\_hold, has\_hold\_f} &
Indicator if a legal hold was active at judge time \\

\textit{Address features} &
ZIP code indicators (top 5 + ``other''), homelessness flag, invalid/missing address flag \\

\bottomrule
\end{tabular}
\caption{Description of engineered features used to model judicial decision-making patterns. Past felony cases are denoted with `\_f` suffix, while misdemeanors have no suffix.}
\label{tab:features}
\end{table*}

In this study, we focused on the years 2020 to 2025, a time period in which the Consent Decree rules had been adopted and were being used consistently. Our data contain information on virtually all criminal cases filed in Harris County between January 1, 2020, and January 8, 2025, with the exception of a small number of sealed and expunged criminal case records, which were provided by the Harris County Office of County Administration’s Research and Analysis Division (RAD). We observe defendant characteristics (such as race, sex, and age) and case characteristics (such as case filing date, charges filed, booking and release dates, and case disposition) for each misdemeanor and felony case in our data. Also available in the data is information about the bond conditions selected by the magistrate for the subset of cases that we focused on: cases that were not given a General Order Bond and were deemed to be carve-out cases according to the new Harris Local Rule 9. These are the cases that proceeded to a magistrate hearing. We have information regarding the pre-trial bond requests made by the District Attorney and by the Public Defender.  We also have information regarding the magistrate’s ruling, including any written reasons that they supplied.  The target that we use for this project is whether or not a personal bond was granted by the magistrate.

The available dataset contained 531,921 unique cases, out of which we filtered out 170,757 cases to be carve-out cases, and we had bail hearing outcome information for 90,429 cases. We also discovered that there were rows with duplicate case IDs with multiple different magistrates, since a particular case may have had multiple bail hearings in the case of rearrests or bond revocations. In these cases we chose to keep the first instance of the case ID, corresponding to the very first bail decision made by a magistrate. We chose this to ensure our model reflects each magistrate's independent assessment based solely on the original offense details and defendant circumstances at case initiation, without contamination from subsequent developments that could introduce data leakage. As a result, we had 64,803 unique cases to work with.

The next step was to filter a special type of cases, which are off-docket cases. We removed off-docket cases because these represent pre-screened cases where pretrial services had already determined the defendant was eligible for release on a personal bond, and where the magistrate agrees, the magistrate does so on paper, granting the personal bond without a full bail hearing. Since the outcome in these cases is influenced by both pretrial services’ initial screening decisions and the magistrate’s subsequent review, they are not completely representative of that individual magistrate's behavior, and we decided that although such rulings are very important and interesting, since our focus is on magistrate decision making, including these off-docket cases could introduce bias. We used the ‘off\_docket’ as well as the ‘other\_comments’ features to check if a case was off docket, and we narrowed the dataset to 30,159 cases.

We then engineered features using the raw data. We combined the crime case details and bond analytics information from both the felony dataset and misdemeanor dataset, for individual defendants to compute the features as shown in Table \ref{tab:features}. These features mostly pertain to that defendant’s criminal history, and each one has two versions: one for past misdemeanors and one for past felonies (with the `\_f` suffix). Felony crime information is included because, after looking at some PC forms, we discovered that magistrates sometimes grant personal bonds despite pending felony charges, reasoning that the ongoing case would ensure punishment if reoffense occurred, making it important for modeling their decisions.

\section{Case Distribution by Judge}
\label{app:CaseDistribution}
\begin{figure}[h]
\centering
\includegraphics[width=0.8\linewidth]{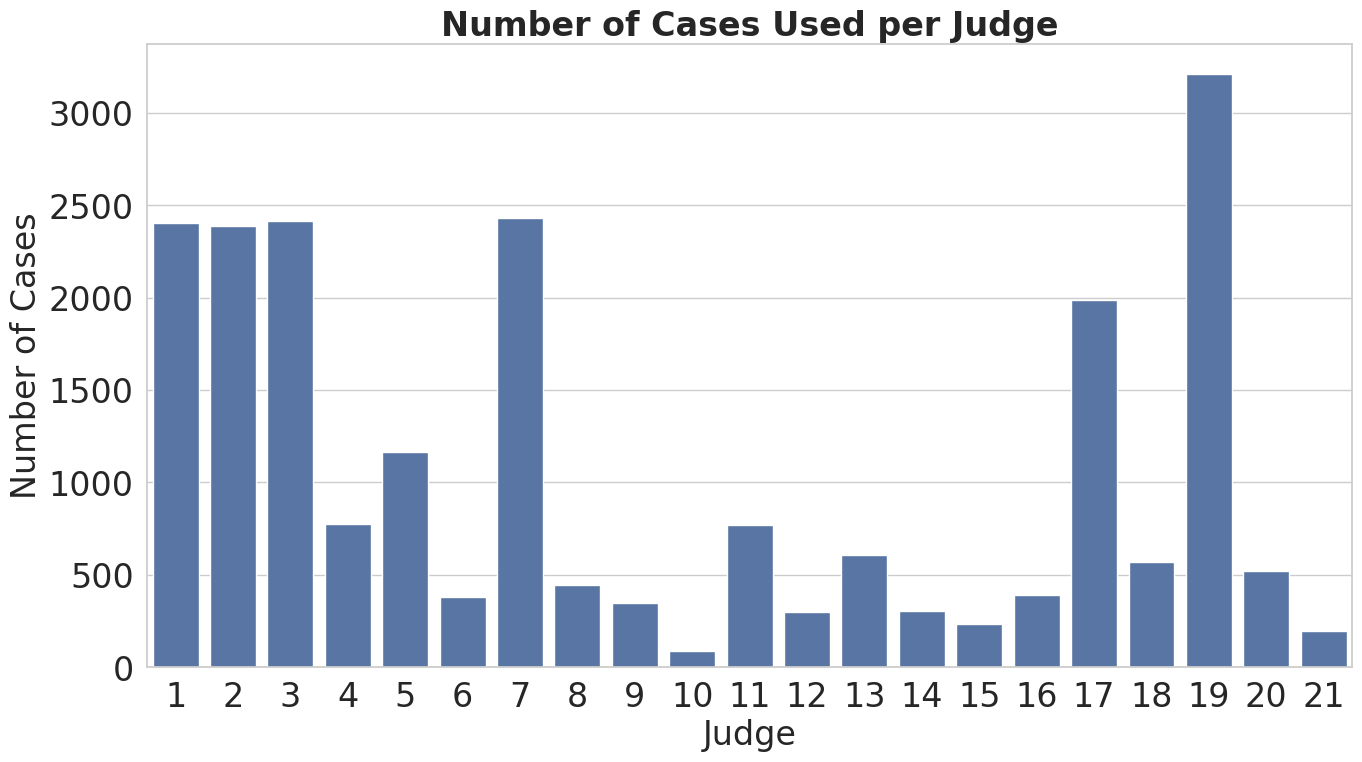}
\caption{Number of Cases for each Judge in the Final Processed Dataset.}
\end{figure}

\begin{figure}[h]
\centering
\includegraphics[width=0.9\linewidth]{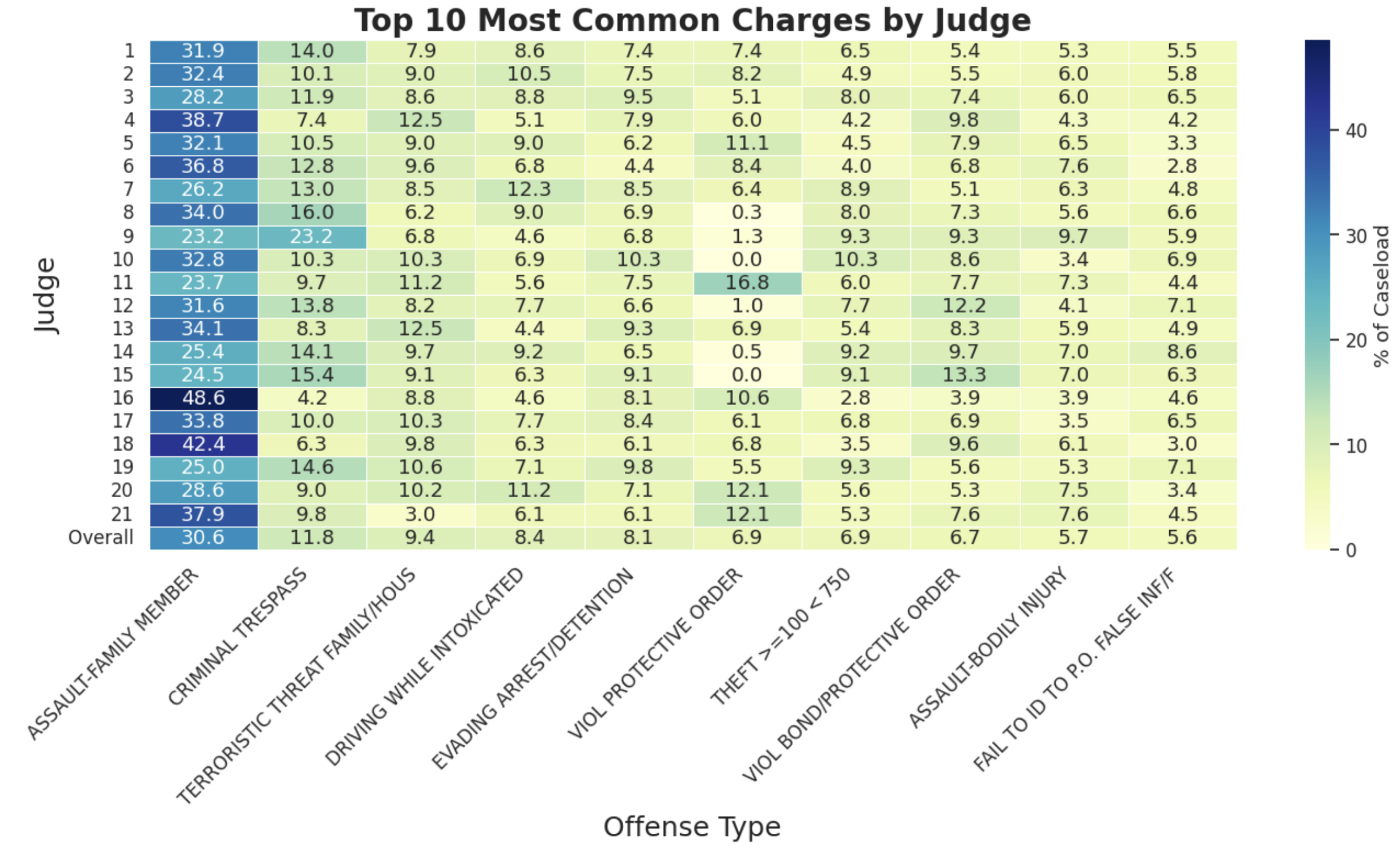}
\caption{Distribution of the Top 10 Most Common Offenses for each Judge.
\label{fig:ChargeDistribution}}
\end{figure}

\begin{figure}[h]
\centering
\includegraphics[width=.9\linewidth]{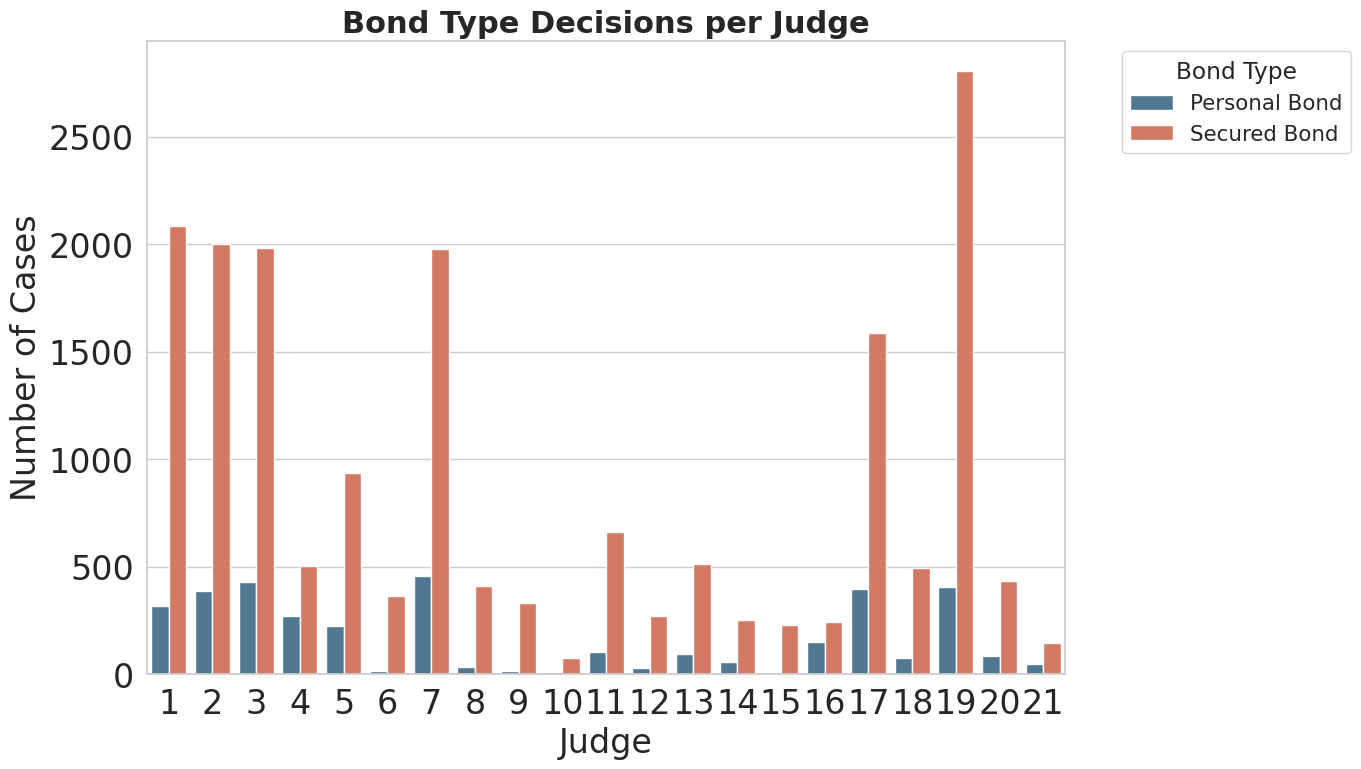}
\caption{Distribution of Bond Type (Personal or Secured) Among the Cases for each Judge.
\label{fig:BondDistribution}}
\end{figure}

Lastly, we also plot the distribution of bond type for each of the judges in our dataset, shown in Figure \ref{fig:BondDistribution}. As we see, the majority of the cases for each judge were granted a secured bond.

% \section{Average Monthly Case Volume Per Judge}
% \label{app:MonthlyDistribution}

% \begin{figure}[ht]
% \centering
% \includegraphics[width=0.8\linewidth]{Figures/MonthlyDistribution.png}
% \Description{Trends in average monthly case volume across specific judges.}
% \caption{Trends in Average Monthly Case Volume Across Specific Judges. Overall for 7 of the judges we see variation throughout the year, with the largest number of cases March and April, but for the rest the average number of cases seems uniform throughout the year. There also seems to be a peak around September to October for some.
% \label{fig:MonthlyDistribution}}
% \end{figure}

\section{Details on Generating the Rashomon Sets}
\label{app:Rashomon}

All the features were transformed using a Threshold Guessing Binarizer, which generates binary splits of numeric features by recursively estimating thresholds via ensembles of shallow depth-1 trees. This step converts numerical columns into interpretable binary indicators such as ``feature $\leq$ threshold,'' which are well-suited for decision tree induction. Each observation’s caseid was preserved for traceability but excluded from modeling. TreeFARMS was then trained with regularization 0.005, a maximum depth of 5, a Rashomon bound multiplier of 0.01, and balanced training enabled. The bound multiplier (0.01) defines the tolerance for including models in the Rashomon set: trees within 1\% of the optimal objective are retained. The trained sets were evaluated on the original data, and the accuracy scores were reported per judge. 

\section{Rashomon Set Samples for Judge 1}
\begin{figure}[h]
    \centering
    \includegraphics[width=\columnwidth]{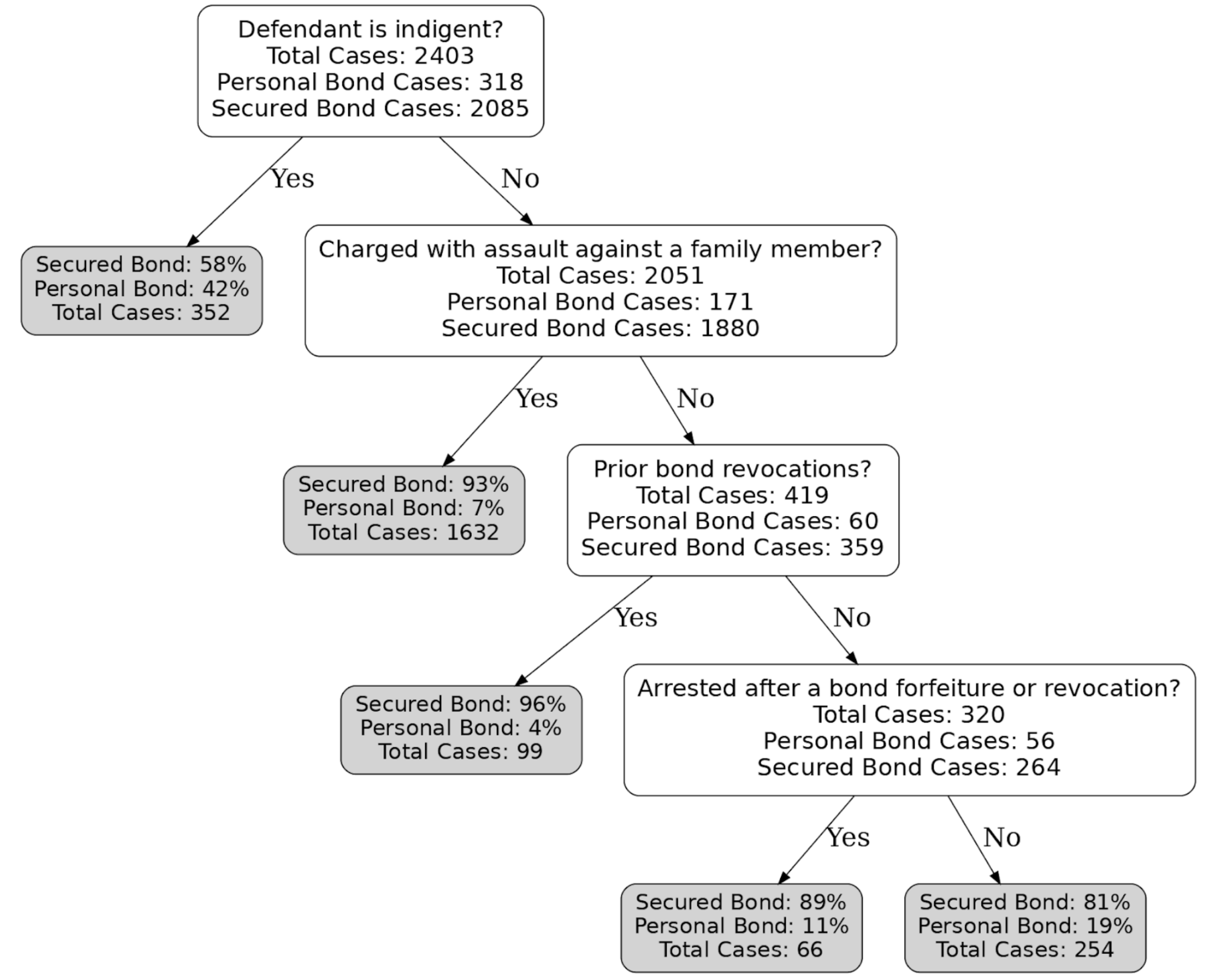}
    \includegraphics[width=\columnwidth]{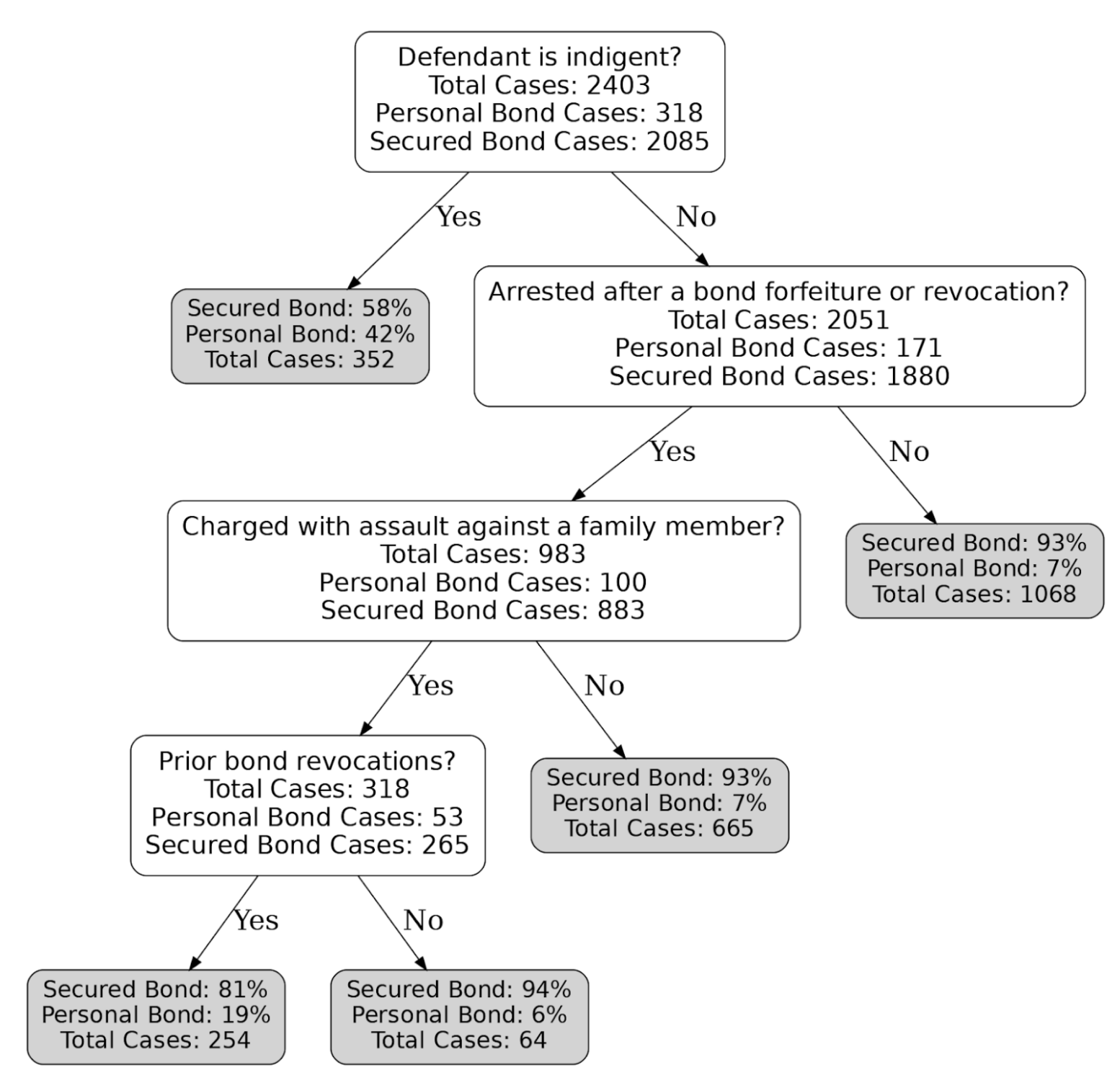}
    \caption{Sample trees from the Rashomon Set of Judge 1.
    \label{fig:Judge1Rashomon}}
\end{figure}

In this section, we look at samples of the Rashomon set of trees for Judge 1. We have more examples of of trees from the Rashomon Set for Judge 1 in Figure \ref{fig:Judge1Rashomon}. In this figure, we see that even though there are many trees that are a part of the Rashomon set, meaning they all are good predictive models for Judge 1, they largely use similar variables. This somewhat shows the consistency of judges, as even over multiple different trainings, they still use similar variables. Furthermore, we see that these trees are still fairly sparse, having a small number of leaves, while averaging high accuracies, showing that judges can largely be represented by simple rules. We saw these types of shallow trees for many of the judges.
\label{app:Judge1}
% \begin{figure}[h]
%     \centering
%     \includegraphics[width=\columnwidth]{Figures/RashomonJudge1Tree1.png}
%     \caption{Sample tree from the Rashomon Set of Judge 1.
% \end{figure}

\section{Rashomon Set Samples for Judge 17}
\begin{figure}[h]
    \centering
    \includegraphics[width=\columnwidth]{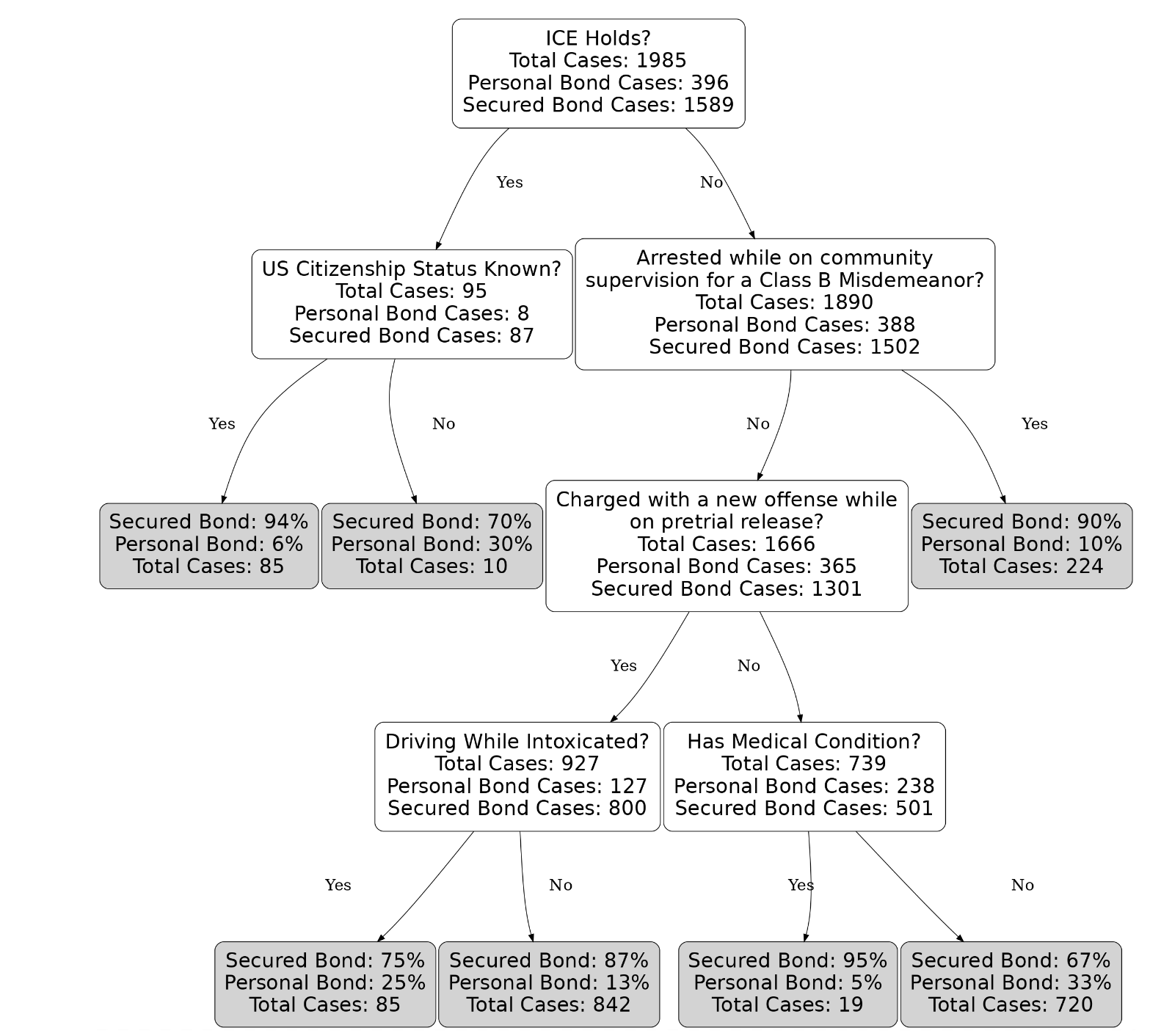}
        \includegraphics[width=\columnwidth]{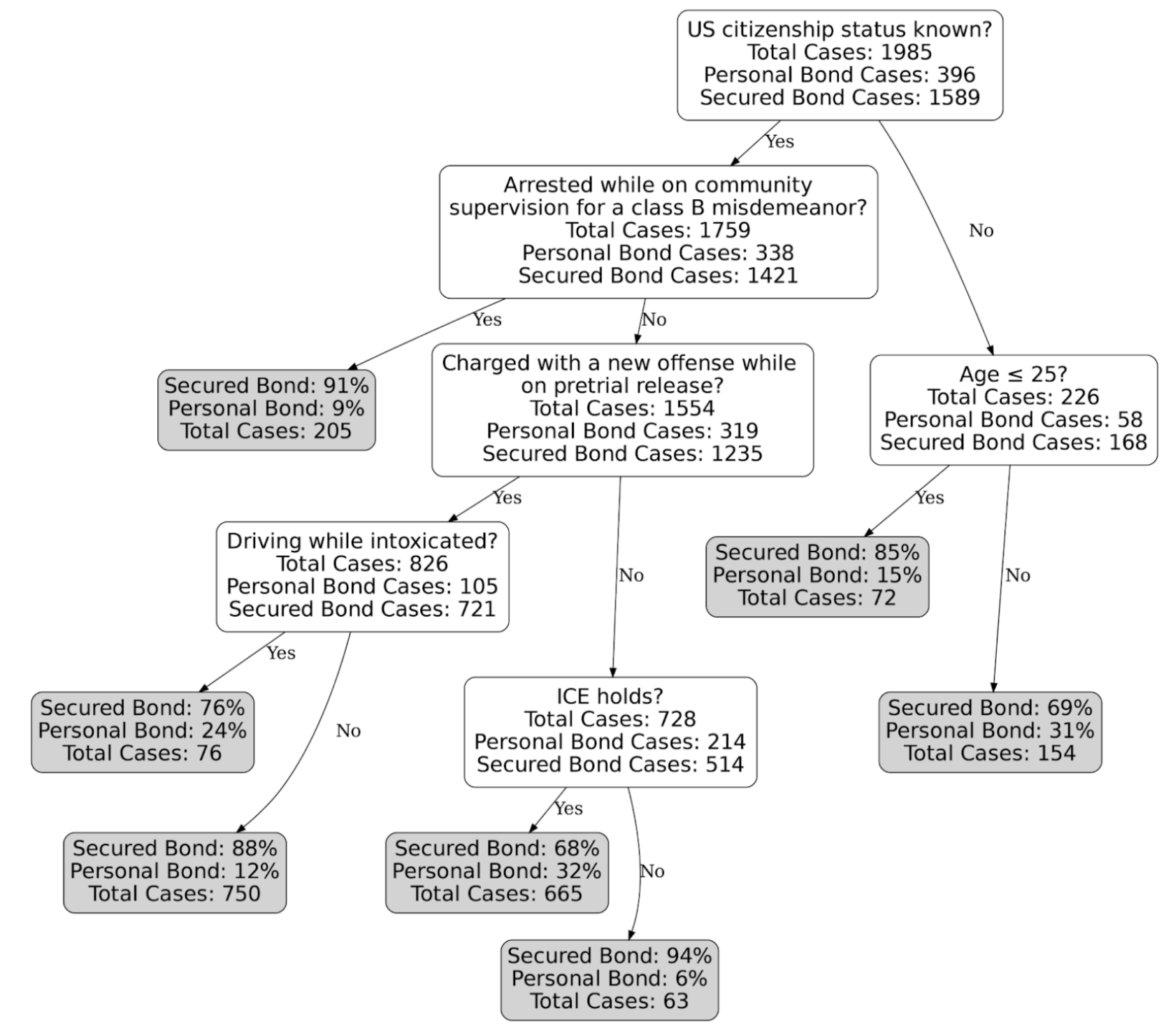}
    \caption{More sample trees from the Rashomon Set of Judge 17.
    \label{fig:Judge17Rashomon}}
\end{figure}

In this section,we focus on Judge 17 specifically. We see in Figure \ref{fig:Judge17Rashomon}, that the Rashomon set can contain trees that use different variables. Even though both of these models are good models for Judge 17, we see that the tree above uses age as a variable to split on, while the second does not and uses whether or not has a medical condition. While there are still some variables that show up in both the trees, we see that it is possible for judges to have different rules that \textit{may} explain a different subset of cases. Again, we note that this behavior appears for many judges, with many trees using different variables, prompting us to do our variable importance analysis to get a better sense of what variables are generally important for each judge.
\label{app:Judge17}

\section{Case Outcome Distribution After Rashomon Filtering}

\begin{figure}[h]
    \centering
    \includegraphics[width=\columnwidth]{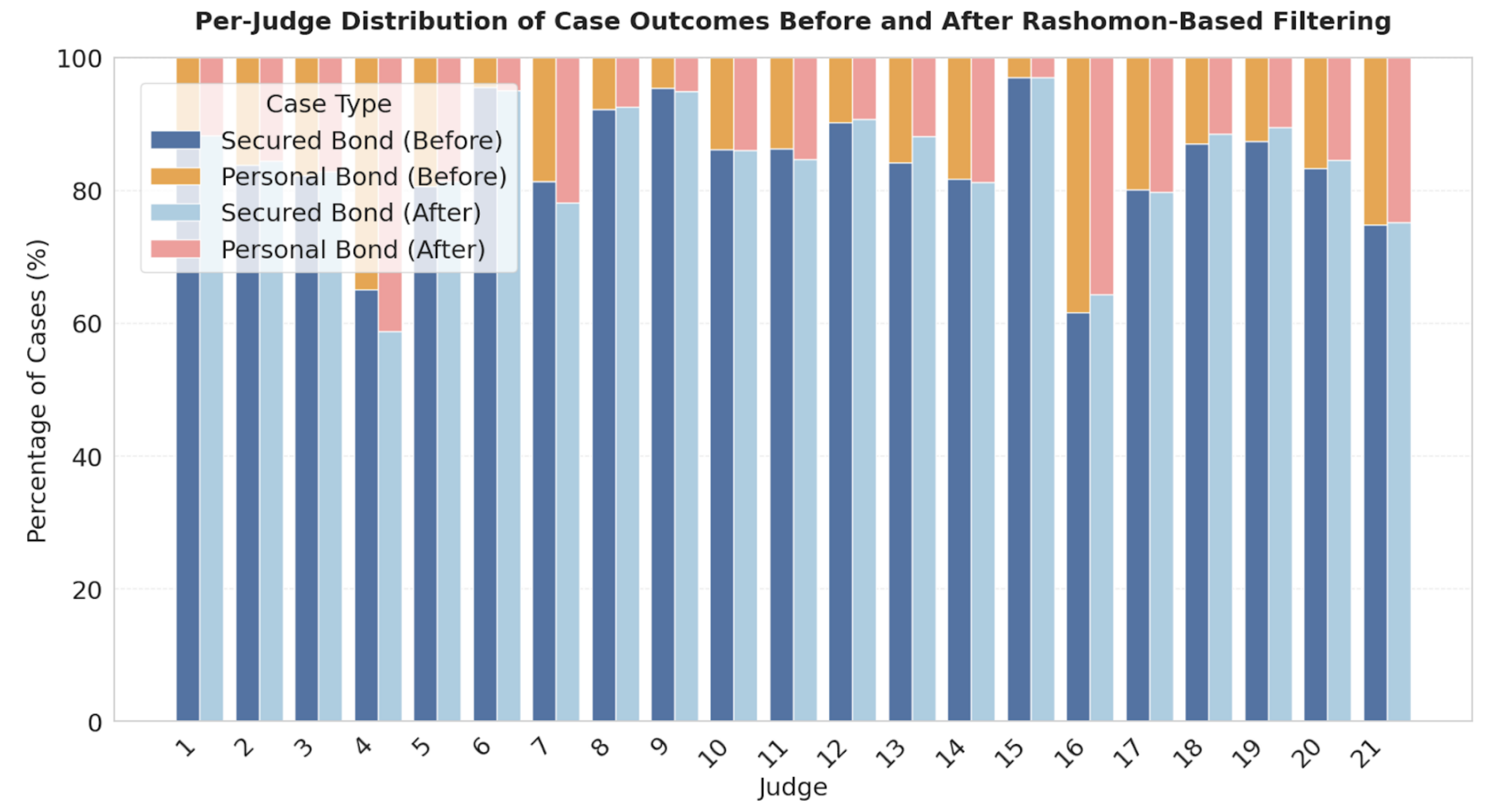}
    \caption{Per-Judge Distribution of Case Outcomes Before and After Rashomon-Based Filtering.}
    \label{fig:filteringdist}
\end{figure}

Here we show the distribution of outcomes of the cases reviewed per judge. In Figure \ref{fig:filteringdist}, we see that after removing the unexplainable cases, there is still a relatively similar ratio of secured bonds and personal bonds granted, implying that no specific class was more unexplinaable. 
\label{app:Filtered}

\section{Do Judges Get Different Cases?}
\label{app:DistDiff}

In our analysis, we wanted to see if judges got a similar distribution of cases, to see if our results are somewhat biased by judges getting different types of cases. First, we trained an XGBoost classifier to see if we could classify the judge that was assigned a case based on the information of the case. If it was truly random, we would expect the classifier to be random, $\sim 4.76\%$ with 21  judges. We saw this classifier get a balanced accuracy of $12.69 \%$, meaning that, while still very bad, there might be \textit{some} difference in the distribution of cases between judges. Therefore, we did two tests, to deal with both the numerical and categorical variables in the data. 

\begin{figure}[h]
    \centering
    \includegraphics[width=\columnwidth]{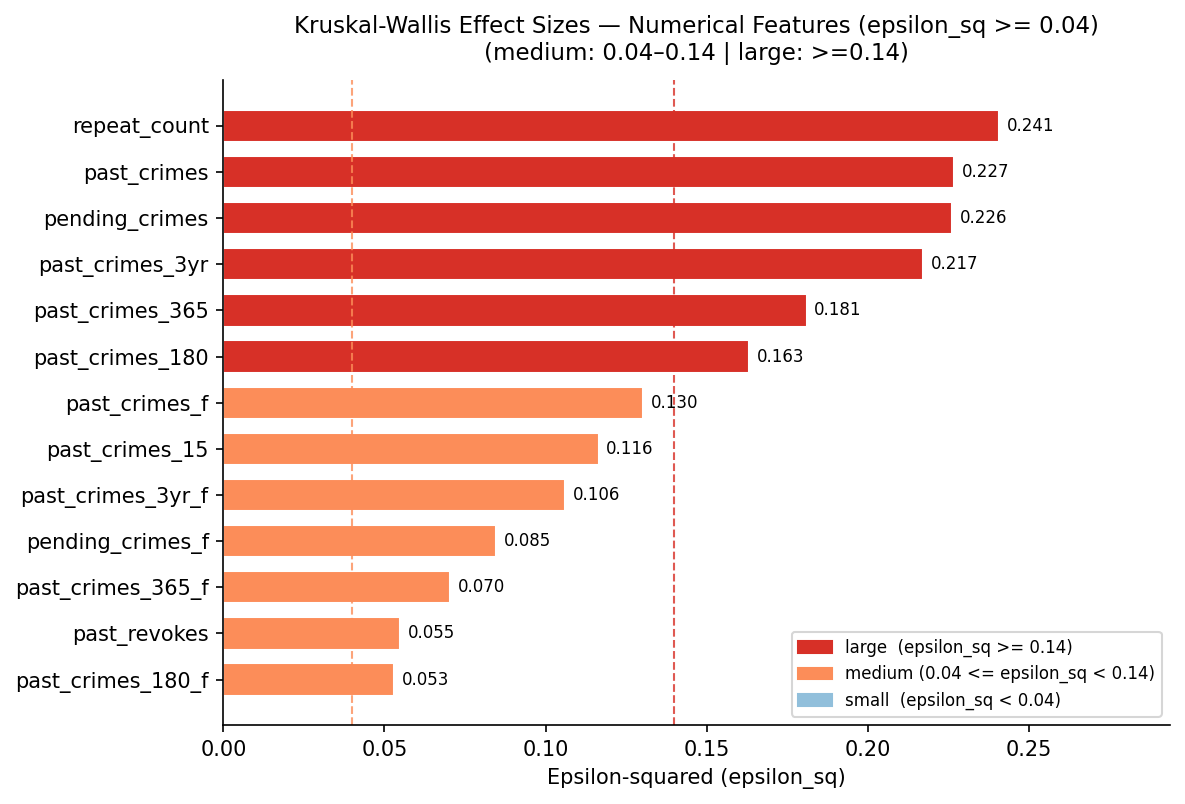}
    \caption{Epsilon Squared effect sizes from the Kurskal-Wallis test on the numerical features that had non-negligible effect sizes.}
    \label{fig:epsilondist}
\end{figure}

First, we did a Kruskal-Wallis test for the numerical variables. We chose to do the Kruskal-Wallis test since our numerical features are integers, not normally distributed, and our judges have a very differnt number of cases. When performing this test, it saw all of these features have a p-value of less than .05, indicating signifigance. Then, we looked at effect sizes, which quantifies how much the variation of the covariate across the dataset was accounted for by which judge a case was assigned to. While some have negligent effect sizes, meaning they really aren't that significant even though they had a low p-value, the majority did have medium to large effect sizes, as seen in Figure \ref{fig:epsilondist}. However, in our variable importance analysis, common splits in the trees, and many of the high confidence, large support rule mining analysis, many of these variables don't actually show up. This seems to indicate that while there may be some differences, potentially due to the large difference in amount of cases, in these variables, they might not matter as much to our analysis, as the very important variables as age and warrant count seem to be consistent between the different judges.

\begin{figure}[h]
    \centering
    \includegraphics[width=\columnwidth]{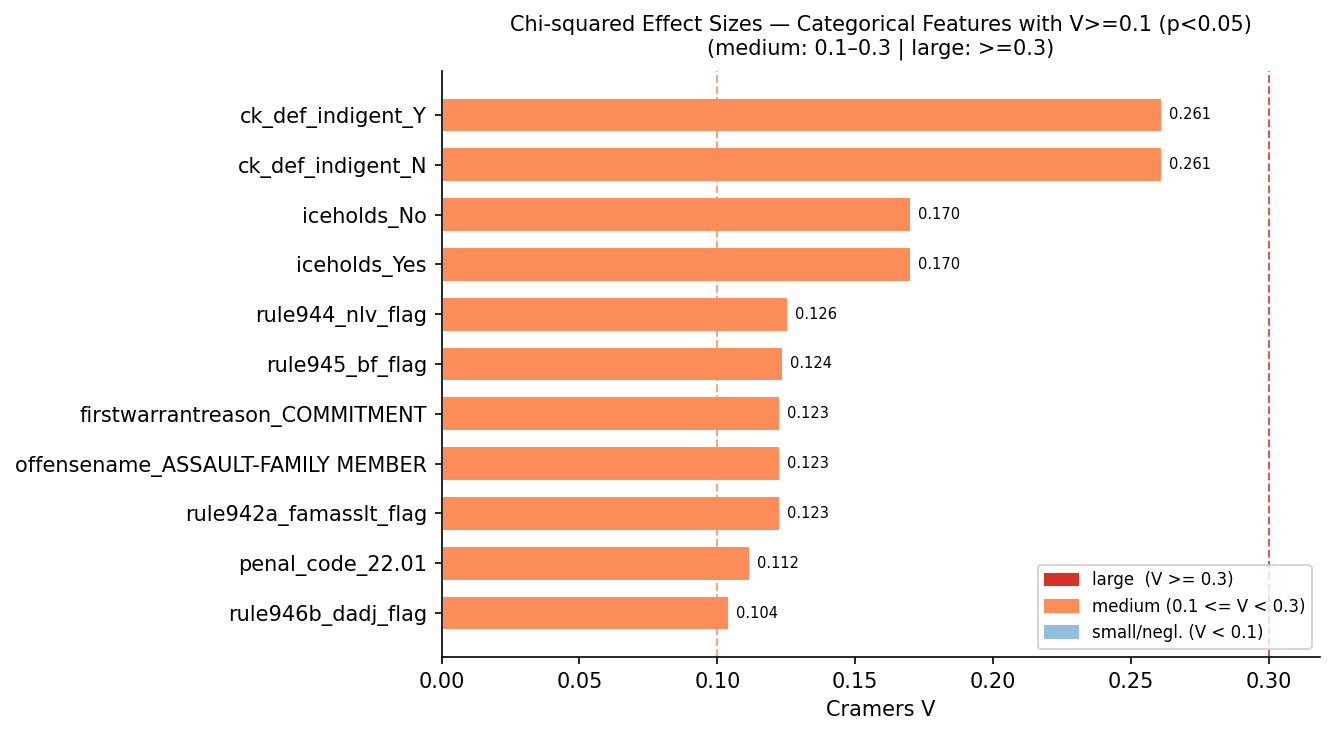}
    \caption{ Cramér's V values for categorical features that had non-negligible effects}
    \label{cramersv}
\end{figure}

Second, we performed Chi-squared tests on the categorical variables to see if the proportion of each feature differ between the judges. After performing this test, we saw that 94 our of our 324 categorical variables saw a p-value lower than .05, which indicates a significant difference. Again, we wanted to observe the effect sizes of all of these "signicantly different" variables, as the chi-squared test statistic is a function of both effect magnitude and sample sizesize, and calculated Cramér's V. Cramér's V is calculated by normalizing the chi-squared statistic by both sample size and table dimensions, reflecting the strength of association between judge assignment and a given covariate. Out of our 94 categorical variables, we saw that 83 of them had negligible effect, and showed the 11 that had a medium to high score in Figure \ref{cramersv}. From this, we concluded that only 11 of our 394 categorical variables had non-negligible effects, meaning that the distributions were relatively similar from these variables. However, in our categorical variables, we see that things like ice holds, whether a defendant is indigent, and specific rules and offenenses are very different. This might explain why some of these are not globally important for all of the judges, through our variable important anlaysis, but also shows that while these distributions may be different, they are all relatively important variables for the judges anyways, indicating that the distributions being different may not have as big of an effect on our analysis. However, it does indicate why most of our large-confidence, high-support disagreement rules use these variables. 

This analysis shows that the there is some difference in case distribution between judges, and may explain some results in our analysis, which is only able to be done because our methods are interpretable, 
\section{GOSDT Accuracies}
In this section, we see that the accuracies mostly follow balanced accuracy. We see that most judges can be explained by reasonable accuracies. This shows that, although the data is imbalanced, the algorithms are not overfitted to predicting the majority class, as the balanced accuracy and accuracy metrics have very similar values for each judge.
\begin{figure}[h]
    \centering
    \includegraphics[width=\columnwidth]{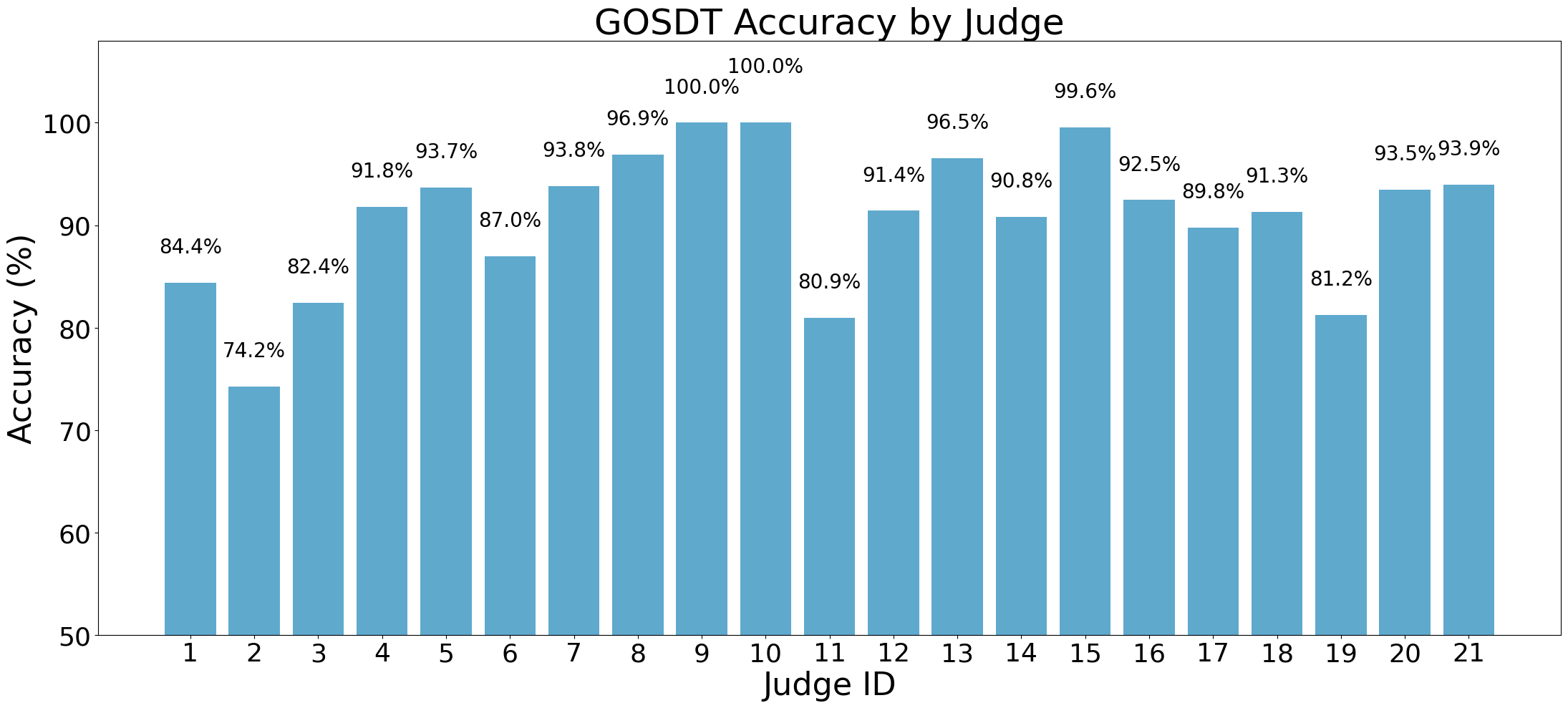}
    \label{GOSDTAccuracy}
    \caption{Accuracy for the GOSDT models trained for each judge on the filtered cases.}

\end{figure}

\section{Example GOSDT trees}
\label{app:gosdt}
\begin{figure}[h]
    \centering
    \includegraphics[width=.9\columnwidth]{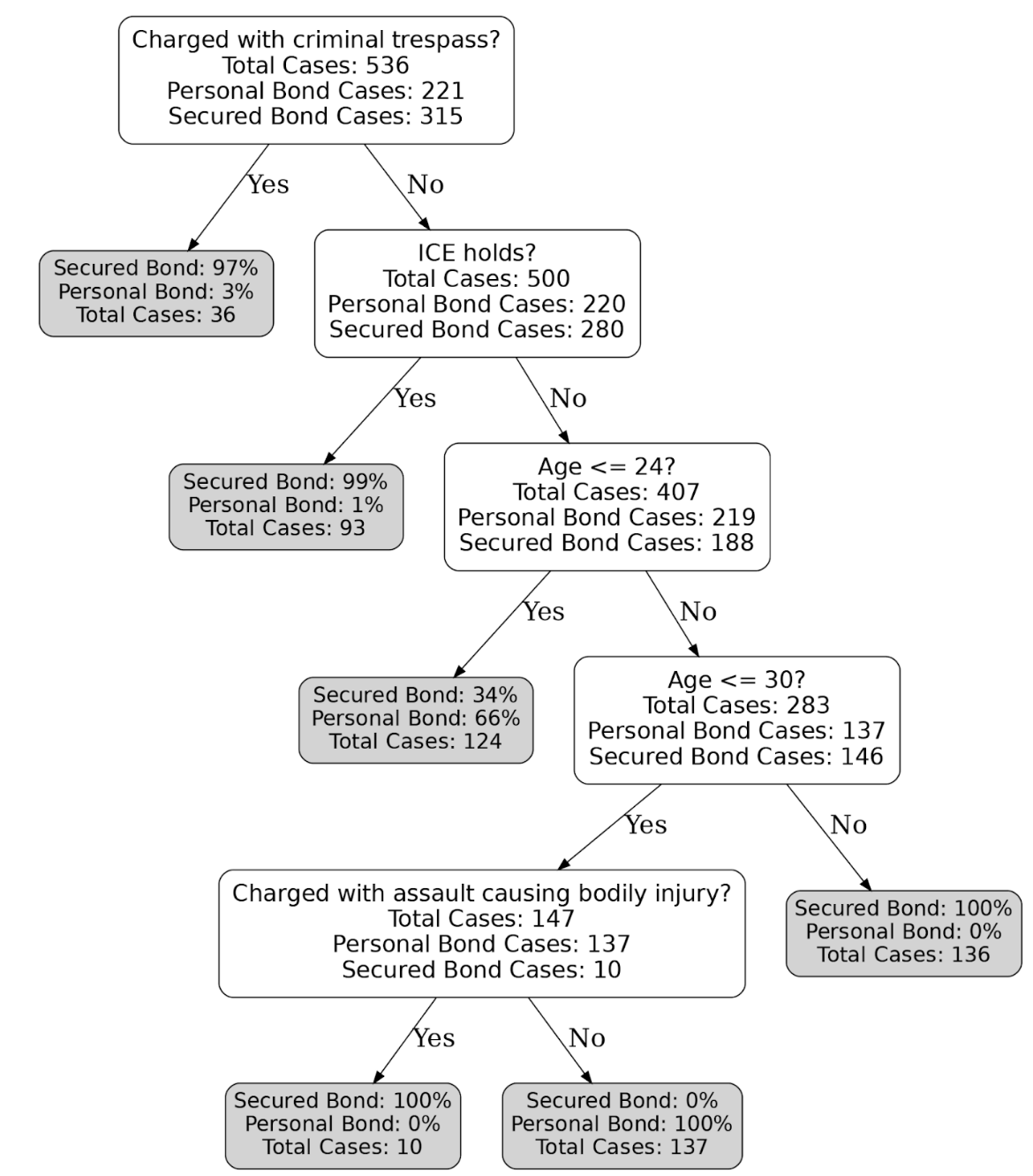}
    \caption{Decision tree for Judge 4 trained using GOSDT.}
    \label{fig:Judge4DT}
\end{figure}

\begin{figure}[h]
    \centering
    \includegraphics[width=.9\linewidth]{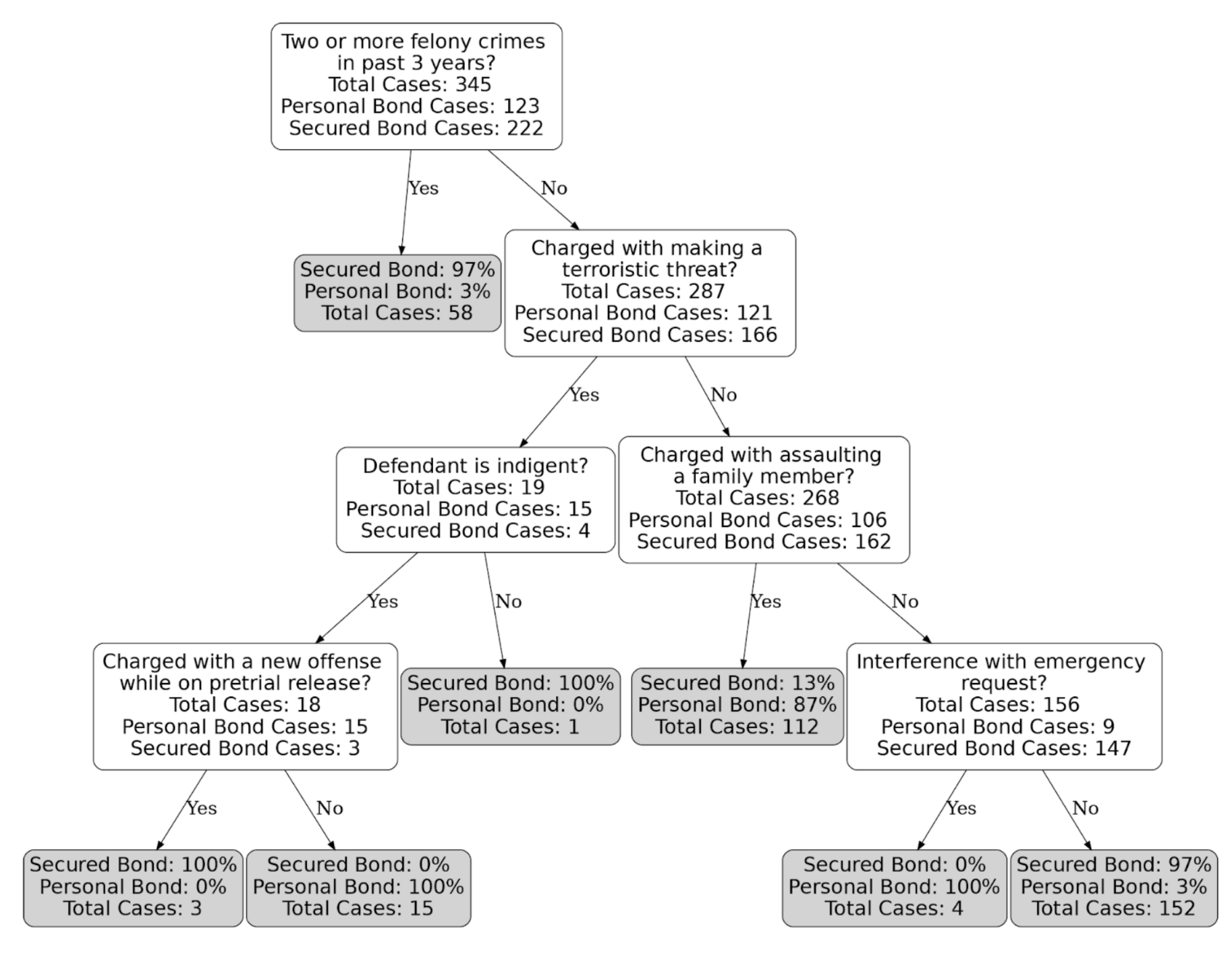}
    \caption{Decision tree for Judge~16 trained using GOSDT.}
    \label{fig:Judge16DT}
\end{figure}

In this section we give some more example of GOSDT trees trained on \textit{only} the remaining cases after we remove the unexplainable cases. We see in Figure \ref{fig:Judge4DT} and Figure \ref{fig:Judge16DT}, that Judges 4 and 16 use very different variables, indicating that judge's individual algorithms may be extremely different.  
\section{Bad GOSDT Tree}
\begin{figure}[h]
    \centering
    \rotatebox{90}{%
        \includegraphics[width=0.95\textheight]{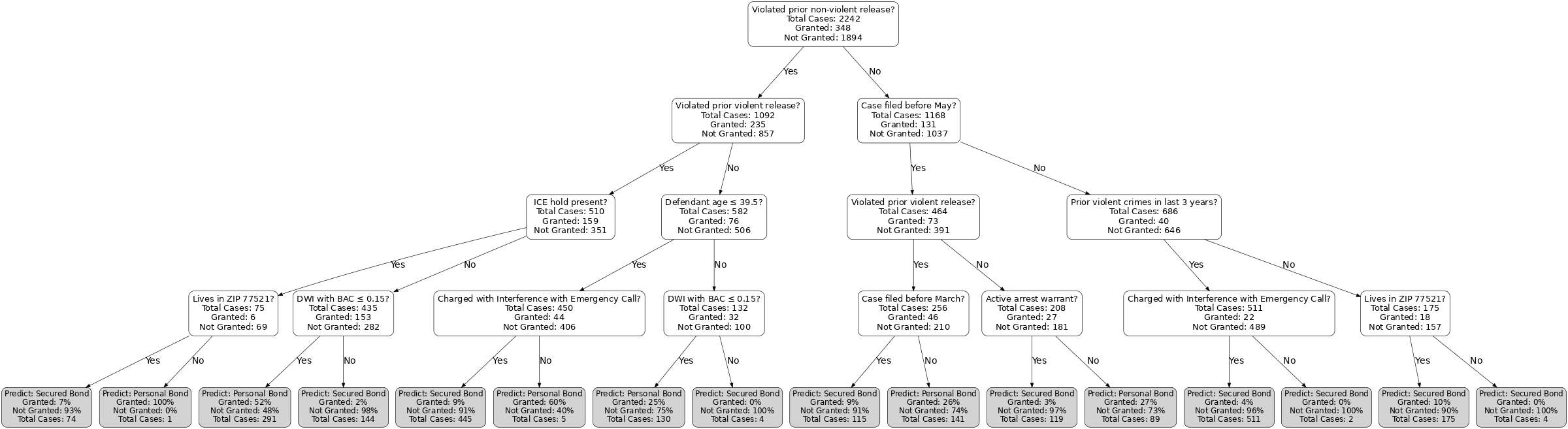}
    }
    \caption{Decision tree learned by GOSDT.}
    \label{fig:BadTree}
\end{figure}

Here, we have an example of a bad GOSDT tree in Figure \ref{fig:BadTree}. We see a full tree that uses every leaf and splits in a lot of different variables. Despite this, and the tree being intentionally overfit as it's very deep and comnplex, we see in the leaves, there are leaves where not many cases are actually predicted correctly. The accuracy is still only about 70\%.
\label{app:BadTree}
\section{Pairwise Loss}
\label{app:loss}
\begin{figure}[ht]
    \centering
    \begin{minipage}[t]{0.5\linewidth}
        \centering
        \includegraphics[width=\linewidth]{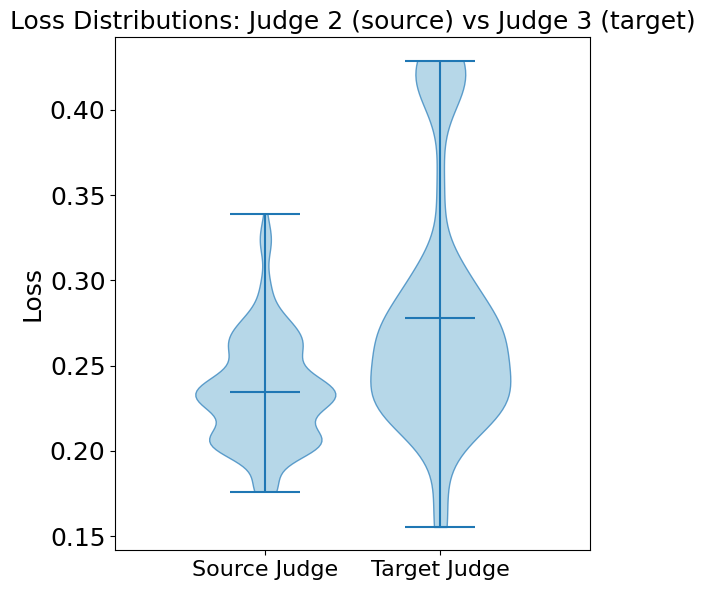}
    \end{minipage}\hfill
    \begin{minipage}[t]{0.5\linewidth}
        \centering
        \includegraphics[width=\linewidth]{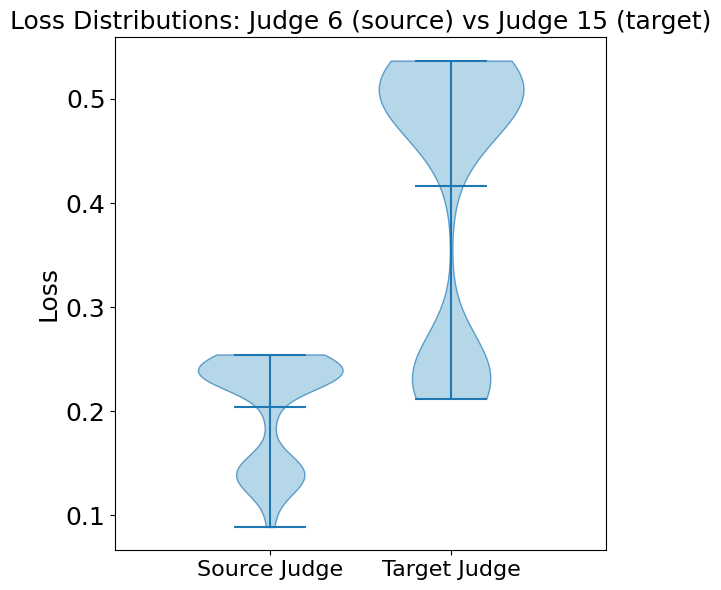}
    \end{minipage}
    \caption{Rashomon model loss distributions on source-judge cases for two additional source--target pairs.
    (\textit{Left}.) Source Judge~2 with target Judge~3.
    (\textit{Right}.) Source Judge~6 with target Judge~15.}
    \label{fig:Loss_other_pairs}
\end{figure}

For most of our pairwise loss analysis, we see the source models perform consistently well on the source data, and sharp rises in and a larger range and more errors when the source models are applied to the target judges' data. However, we also see two other types of loss distributions. In Figure \ref{fig:Loss_other_pairs} (left), the Rashomon models of Judge 2 maintain comparable performance when evaluated on Judge 3, with overlapping distributions and similar median error rates of 0.24 to 0.27. This generally occurs when the loss for the source models for the source judge is not the best, as indicated by the loss for the source judge in Figure \ref{fig:Loss_other_pairs} (left) being much higher than the loss when a model is accurate. In Figure \ref{fig:Loss_other_pairs} (right), we see another type of comparison. Some models from the source set transfer well, as we see some overlap in loss distributions. This indicates that, sometimes, judges can have subsets of cases their algorithms agree on, which may indicate some agreement across judges.

\end{document}